\documentclass[conference]{IEEEtran}
\usepackage{times}

\usepackage[utf8]{inputenc}
\usepackage[T1]{fontenc}
\usepackage[english]{babel}  % Language hyphenation and typographical rules
\usepackage{mathtools}  % Improved amsmath
\usepackage[bb=dsserif]{mathalpha}  % used for blackboard numbers
\usepackage{bm}
\usepackage{amsfonts}
\usepackage{float}  % Improved interface for floating objects
\usepackage[final, colorlinks = true]{hyperref}  % For hyperlinks in the PDF
\usepackage{graphicx}  % for pdf, bitmapped graphics files
\usepackage{subcaption}  % Support for sub-figures
\usepackage{booktabs}  % Better looking tables
\usepackage{csquotes}  % Context sensitive quotation facilities
\usepackage{cleveref}
\usepackage{microtype}  % Slightly tweak font spacing for aesthetics
\usepackage{siunitx}
\usepackage{ifthen}
\usepackage{balance}
\usepackage[yyyymmdd]{datetime}  % Uses YEAR-MONTH-DAY format for dates
\usepackage{pseudocode}
\usepackage[linesnumbered, vlined, ruled]{algorithm2e}

\newlength{\twosubht}
\newsavebox{\twosubbox}

\usepackage[
backend=biber,
style=numeric,
sorting=nty,
mincitenames=1,
maxcitenames=3,
giveninits=true,
url=false,
]{biblatex}
\newcommand*{\field}[1]{\mathbb{\MakeUppercase{#1}}}  % scalar field
\newcommand*{\set}[1]{{\mathcal{\MakeUppercase{#1}}}}  % set symbol
\newcommand*{\R}{\field{R}} % field of real numbers, or the real line
\newcommand*{\argmax}{\operatornamewithlimits{argmax}}
\newcommand*{\argmin}{\operatornamewithlimits{argmin}}

\newcommand*{\seq}[1]{\MakeUppercase{#1}}

\newcommand*{\diff}{{\operatorname{d}}}

\newcommand*{\given}{|}

\newcommand\Bbbone{%
  \ifdefined\mathbbb%
    \mathbbb{1}%
  \else%
    \boldsymbol{\mathbb{1}}%
  \fi}
 
\renewcommand{\vec}[1]{{\boldsymbol{\mathbf{#1}}}}
\newcommand*{\mat}[1]{{\MakeUppercase{#1}}}
\newcommand*{\eye}{\mat{I}}							% the identity matrix
\newcommand*{\transpose}{\mathsf{T}}
\newcommand*{\pDensity}{p}

\newcommand*{\expectation}{\mathbb{E}}

\newcommand*{\kl}[2]{D_\mathrm{KL}(#1||#2)}
\newcommand*{\normal}{\mathcal{N}}					% normal distribution
\newcommand*{\dirac}[1]{\delta(#1)}

\newcommand*{\rv}[1]{\MakeUppercase{#1}}	% random variable
\newcommand*{\indep}{\perp \!\!\! \perp} % independent

\newcommand*{\qClass}{\set{Q}}

\newcommand*{\qDensity}{q}

\newcommand*{\primIdx}{i}	% primary index
\newcommand*{\secIdx}{j}	% secondary index

\newcommand*{\tIdx}{t}

\newcommand*{\state}{\vec{x}}
\newcommand*{\stateSeq}{\seq{x}}
\newcommand*{\stateSpace}{\set{X}}

\newcommand*{\trajectory}{\vec{\tau}}
\newcommand*{\trajectoryIdx}{n}
\newcommand*{\trajectorySamples}{N_{\trajectory}}
\newcommand*{\trajectorySpace}{\set{T}}

\newcommand*{\seqOp}{\mathcal{H}}

\newcommand*{\control}{\vec{u}}
\newcommand*{\controlSeq}{\seq{u}}
\newcommand*{\controlSpace}{\set{U}}

\newcommand*{\instCost}{c}
\newcommand*{\termCost}{c_{term}}
\newcommand*{\cost}{C}

\newcommand*{\optimality}{\mathcal{O}}

\newcommand*{\controlHorizon}{H}
\newcommand*{\hzIdx}{h}

\newcommand*{\weight}{\omega}

\newcommand*{\policy}{\pi}
\newcommand*{\polParams}{\vec{\theta}}
\newcommand*{\polIdx}{i}

\newcommand*{\polSpace}{\Theta}
\newcommand*{\polSize}{{N_\policy}}  % Number of Stein particles
\newcommand*{\polSamples}{N_{\operatorname{a}}}
\newcommand*{\polPriorCov}{\mat{\Sigma}_{\operatorname{a}}}
\newcommand*{\controlAuth}{\mat{\Sigma}}

\newcommand*{\simParams}{\vec{\xi}} % parameters for a single input location's distribution
\newcommand*{\simSpace}{\Xi} % set of parameters, each for a single input location
\newcommand*{\simFunction}{\hat{f}_{\simParams}}
\newcommand*{\simDensity}{p_\simParams}
\newcommand*{\simIdx}{m} % set of parameters, each for a single input location
\newcommand*{\simSize}{{N_\simParams}}  % Number of Stein particles
\newcommand*{\simSamples}{N_{\operatorname{s}}}
\newcommand*{\system}{f}

\newcommand*{\partIdx}{i}
\newcommand*{\partSize}{N_p}
\newcommand*{\partDim}{p}
\newcommand*{\scoreFunc}{\vec{\phi}}
\newcommand*{\stepSize}{\epsilon}
\newcommand*{\likelihood}{\ell}
\newcommand*{\steinRKHS}{\mathcal{S}}

\newcommand*{\spread}{\alpha}

\newcommand*{\xicr}{x_{{\operatorname{ICR}}}}
\newcommand*{\wradius}{r_{{\operatorname{w}}}}
\newcommand*{\axdist}{a_{{\operatorname{w}}}}

\newcommand*{\mpfLikCov}{\mat{\Sigma}_{\operatorname{obs}}}
\newcommand*{\mpfPriorCov}{\mat{\Sigma}_{\operatorname{s}}}
\newcommand*{\mpfSteps}{L}

\newcommand*{\datapoint}[1][\tIdx]{\mathcal{D}_{#1}}
\newcommand*{\dataset}[1][\tIdx]{\mathcal{D}_{1:#1}}
\newcommand*{\dataSize}{N_\mathcal{D}}

\newcommand*{\noise}{\rv{\eta}}	% noise random variable

\newcommand*{\anyvector}{\vec{x}}
\newcommand*{\anyerror}{\vec{e}}

\newcommand*{\iid}{i.i.d.\xspace}
\newcommand*{\wrt}{w.r.t.\xspace}
\newcommand*{\ie}{i.e.\xspace}
\newcommand*{\eg}{e.g.\xspace}
\newcommand*{\st}{s.t.\xspace}

\newcommand*{\dust}{DuSt-MPC\xspace}
\newcommand*{\disco}{DISCO\xspace}
\newcommand*{\sv}{SVMPC\xspace}

\title{Dual Online Stein Variational Inference \\
for Control and Dynamics}

\author{
\authorblockN{
Lucas Barcelos\authorrefmark{1},
Alexander Lambert\authorrefmark{2},
Rafael Oliveira\authorrefmark{1},
Paulo Borges\authorrefmark{3},
Byron Boots\authorrefmark{4}\authorrefmark{5},
and
Fabio Ramos\authorrefmark{1}\authorrefmark{5}% <-this % stops a space
}
\authorblockA{
\authorrefmark{1}The University of Sydney, 
\authorrefmark{2}Georgia Institute of Technology, 
\authorrefmark{3}CSIRO, 
\authorrefmark{4}University of Washington, 
\authorrefmark{5}NVIDIA
}
}

\begin{document}
\maketitle
\IEEEpeerreviewmaketitle

%===============================================================================
% \section{Abstract}
%===============================================================================
\begin{abstract}
Model predictive control (MPC) schemes have a proven track record for delivering aggressive and robust performance in many challenging control tasks, coping with nonlinear system dynamics, constraints, and observational noise. Despite their success, these methods often rely on simple control distributions, which can limit their performance in highly uncertain and complex environments. MPC frameworks must be able to accommodate changing distributions over system parameters, based on the most recent measurements. In this paper, we devise an implicit variational inference algorithm able to estimate distributions over model parameters and control inputs on-the-fly. The method incorporates Stein Variational gradient descent to approximate the target distributions as a collection of particles, and performs updates based on a Bayesian formulation. This enables the approximation of complex multi-modal posterior distributions, typically occurring in challenging and realistic robot navigation tasks. We demonstrate our approach on both simulated and real-world experiments requiring real-time execution in the face of dynamically changing environments.
\end{abstract}

%===============================================================================
\section{Introduction}
%===============================================================================

% Formulate the problem and contributions. We have shown in SVMPC that using Stein
% helps in problems with intrinsic multi-modal posterior distribution. On the
% other hand, we have shown in DISCO how leveraging model uncertainty can help on
% the control problem.
Real robotics applications are invariably subjected to uncertainty arising from either unknown model parameters or stochastic environments.
To address the robustness of control strategies, many frameworks have been proposed of which model predictive control is one of the most successful and popular~\cite{camacho_model_2013}. 
MPC has become a primary control method for handling nonlinear system dynamics and constraints on input, output and state, taking into account performance criteria. 
It originally gained popularity in chemical and processes control~\cite{eaton1992model}, being more recently adapted to various fields, such agricultural machinery~\cite{ding2018model}, automotive systems~\cite{di2019automotive}, and robotics~\cite{kamel2017model,zanon2014model}.
In its essence, MPC relies on different optimisation strategies to find a sequence of actions over a given control horizon that minimises an optimality criteria defined by a cost function.

Despite their success in practical applications, traditional dynamic-programming approaches to MPC (such as iLQR and DDP~\cite{tassa2014control}) rely on a differentiable cost function and dynamics model. Stochastic Optimal Control variants, such as iLQG~\cite{todorov2005generalized} and PDDP~\cite{pan2014probabilistic}, can accommodate stochastic dynamics, but only under simplifying assumptions such as additive Gaussian noise. These approaches are generally less effective in addressing complex distributions over actions, and it is unclear how these methods should incorporate model uncertainty, if any.
In contrast, sampling-based control schemes have gained increasing popularity for their general robustness to model uncertainty, ease of implementation, and ability to contend with sparse cost functions~\cite{williams2018robust}.

To address some of these challenges, several approaches have been proposed. 
In sampling based Stochastic Optimal Control (SOC), the optimisation problem is 
replaced by a sampling-based algorithm that tries to approximate an optimal 
distribution over actions~\cite{williamsModelPredictivePath2017}.
This has shown promising results in several applications and addresses some of the former disadvantages, however it may inadequately capture the complexity of the true posterior.
\begin{figure}
    \sbox\twosubbox{
      \resizebox{\dimexpr\linewidth-1em}{!}{
        \includegraphics[height=3cm]{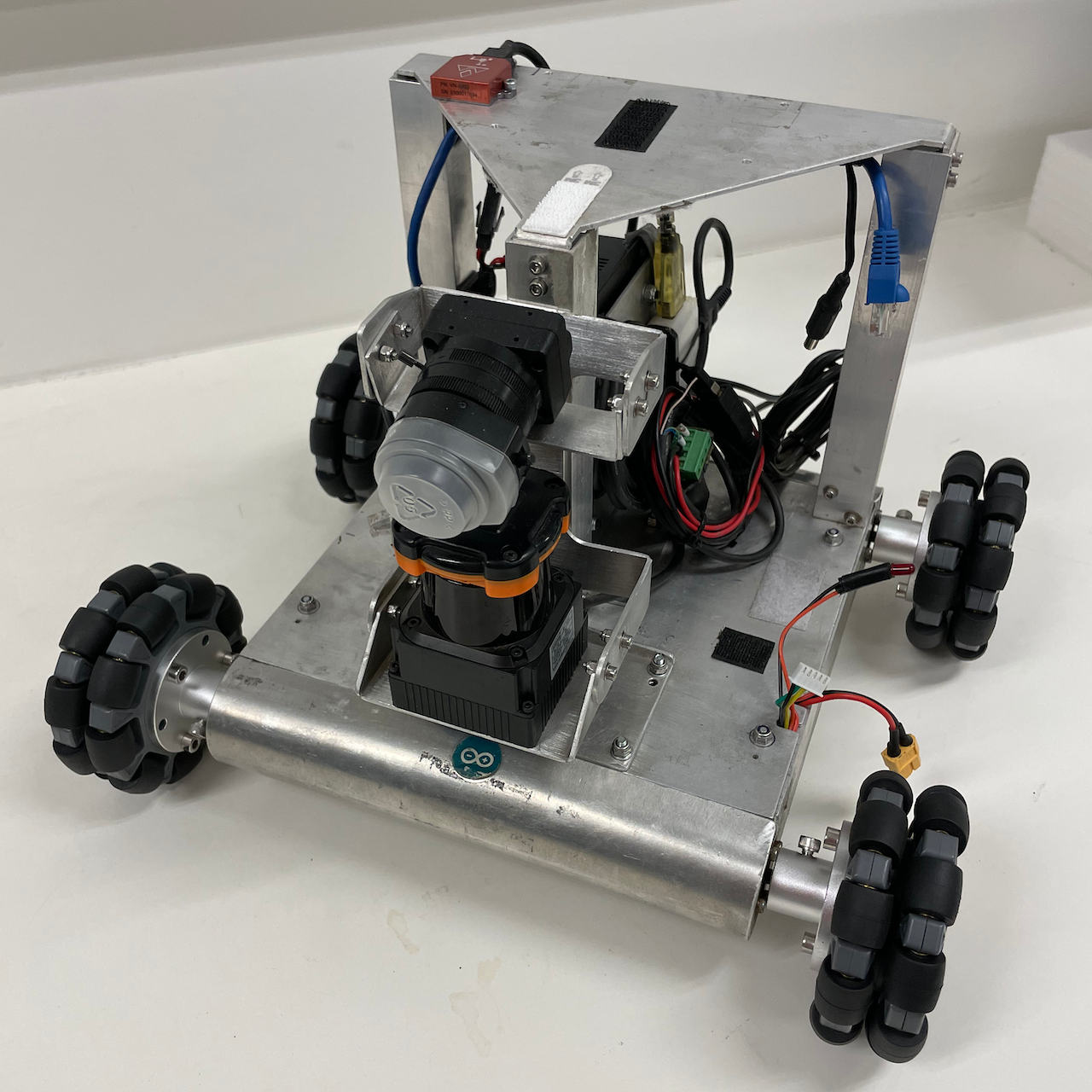}
        \includegraphics[height=3cm]{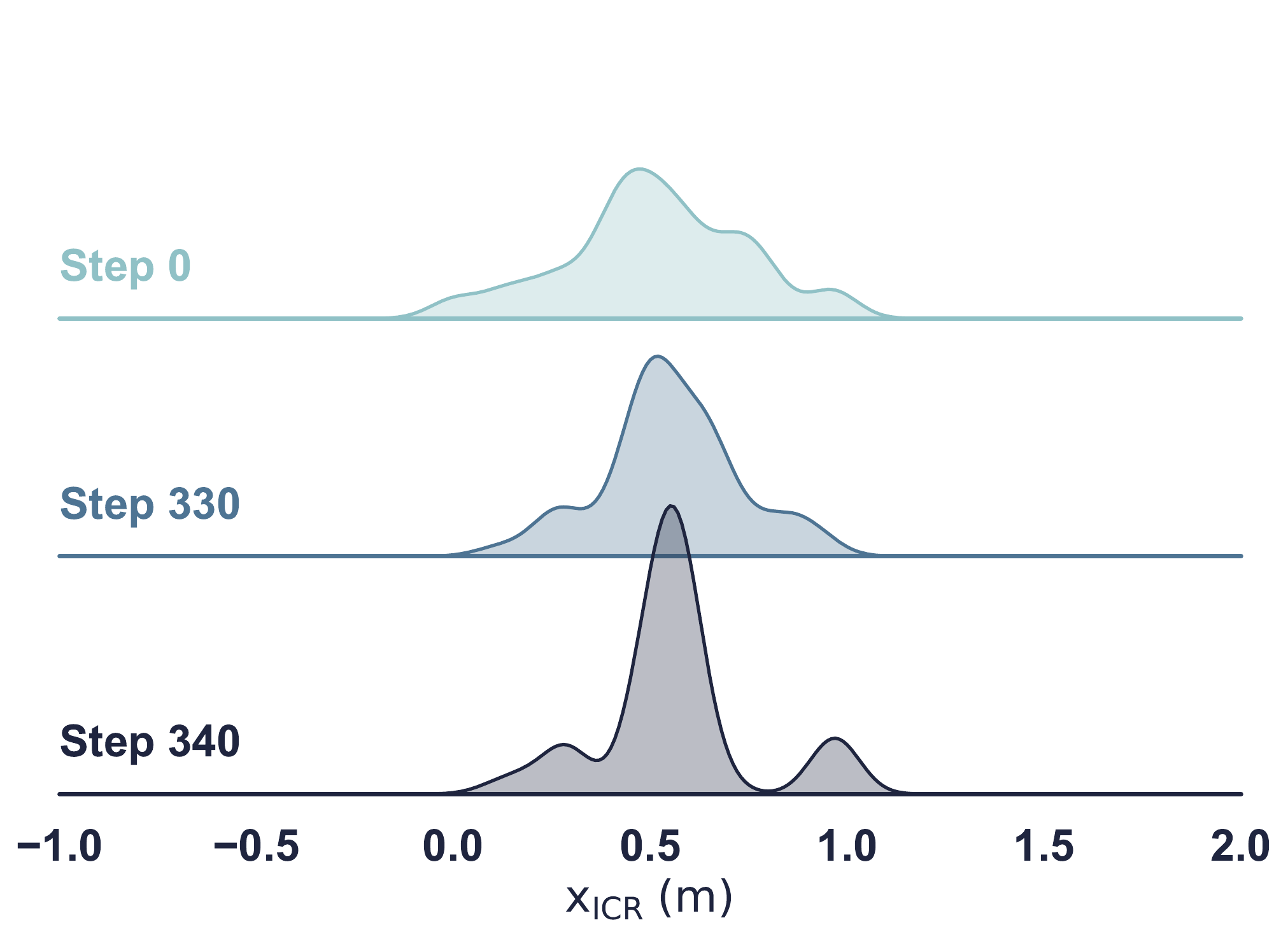}
      }
    }
    \setlength{\twosubht}{\ht\twosubbox}
    \centering
    \subcaptionbox{
    Wombot AGV \label{fig:wombot}
    }{
    \includegraphics[height=\twosubht]{fig/wombot.png}
    }
    \subcaptionbox{
    Posterior distribution over $\xicr$\label{fig:wombot_ridge}
    }{
    \includegraphics[height=\twosubht]{fig/wombot_ridge.pdf}
    }
    \caption{\textbf{Online parameter estimation for autonomous ground vehicles}. Distributions over system parameters such as the inertial center of rotation (ICR), are adapted in real-time.
    (a) The custom built skid-steer robot platform used in experiments.
    (b) Distribution over $\xicr$ at different time steps.
    The mass load on the robot is suddenly increased during system execution. The parameter distribution estimate quickly changes to include a second mode that better
    explains the new dynamics. Our particle-based control scheme can accommodate such multi-modal uncertainty and adapt to dynamically changing environments.}
    \label{fig:motivating}\vspace{-10pt}
\end{figure}

\begin{figure*}
    \centering
    \begin{subfigure}[b]{0.26\textwidth}
    \centering
    \includegraphics[height=4.5cm]{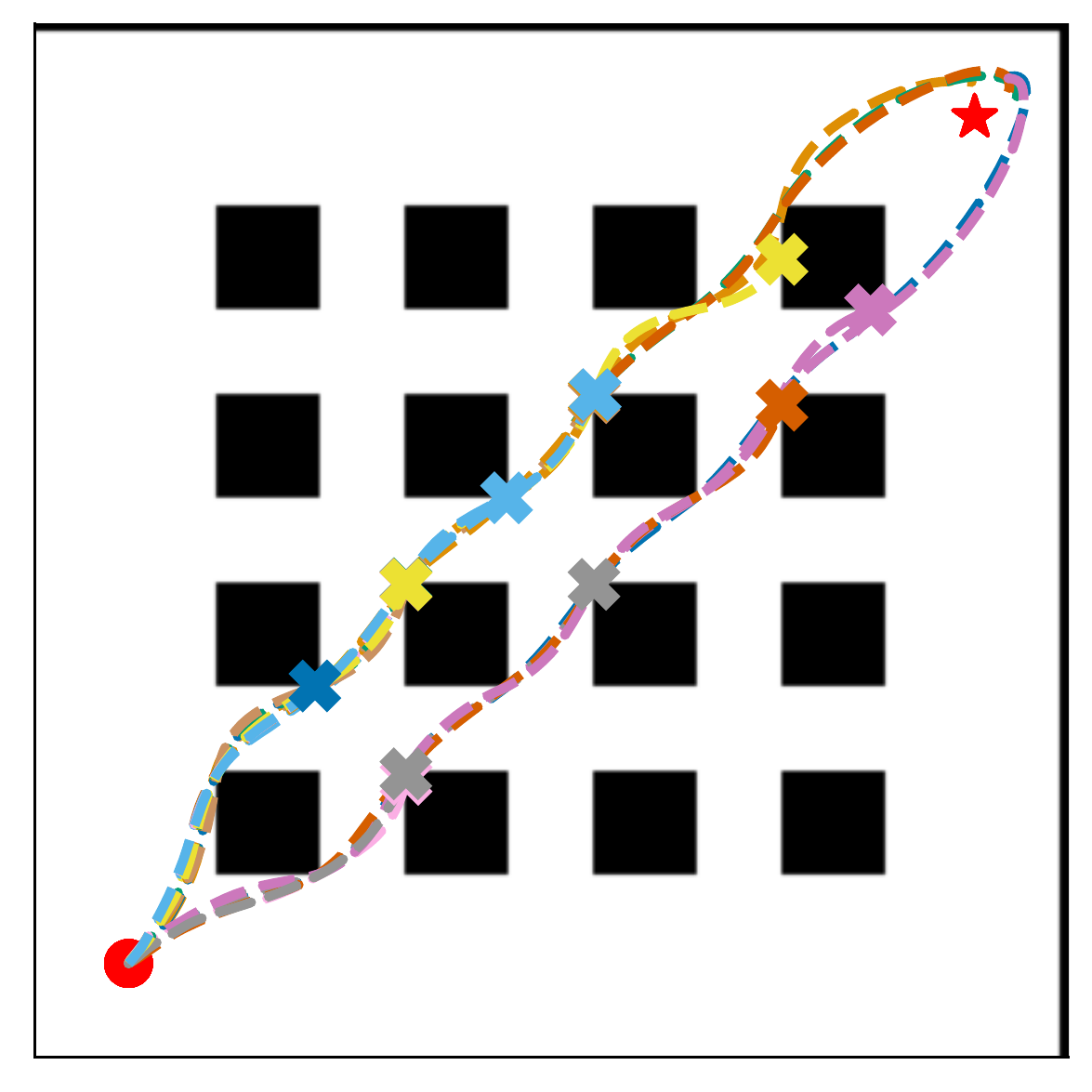}
    \caption{\sv trajectories (baseline)}
    \end{subfigure}
    \begin{subfigure}[b]{0.26\textwidth}
    \centering
    \includegraphics[height=4.5cm]{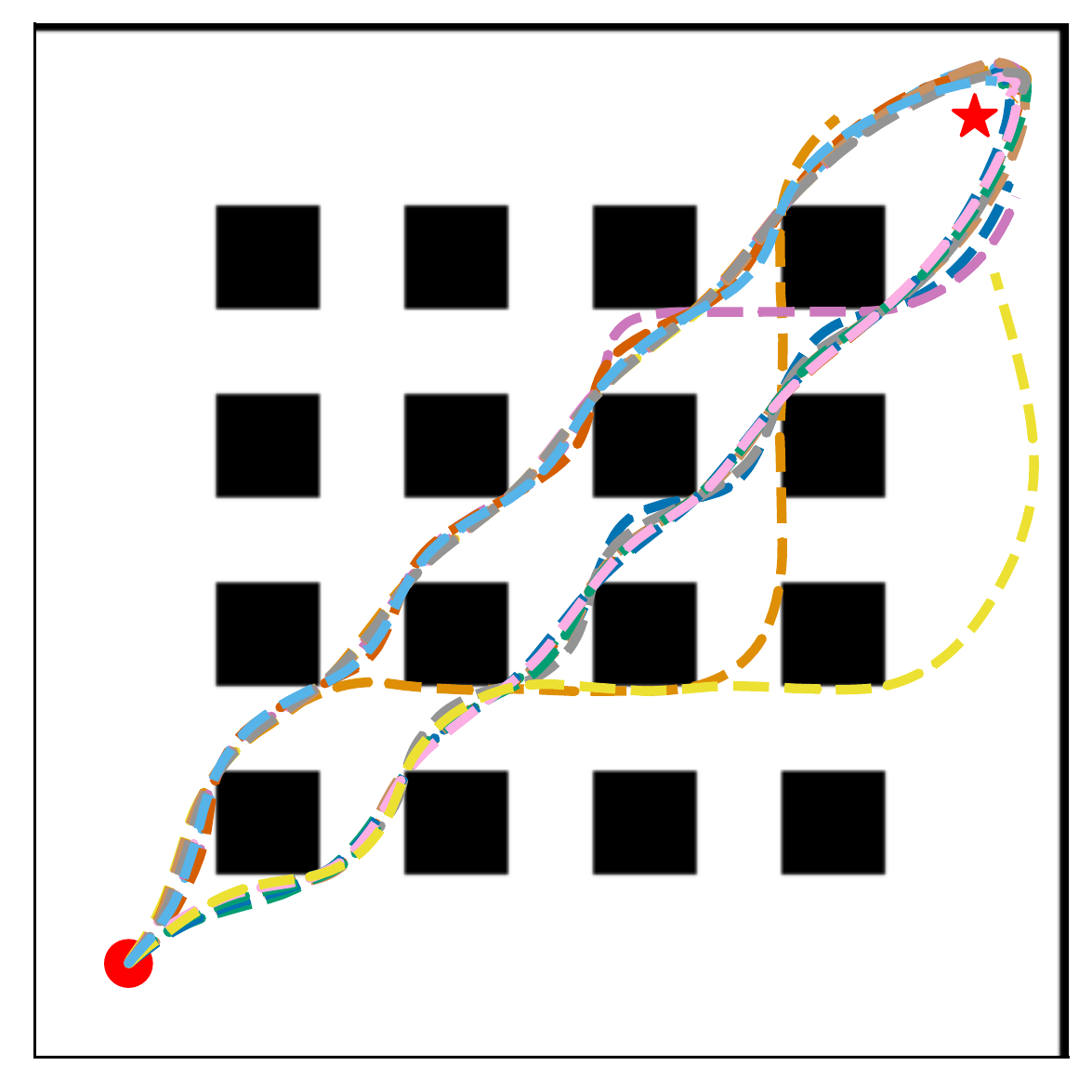}
    \caption{\dust trajectories (ours)}
    \end{subfigure}
    \begin{subfigure}[b]{0.46\textwidth}
    \centering
    \includegraphics[height=4.8cm]{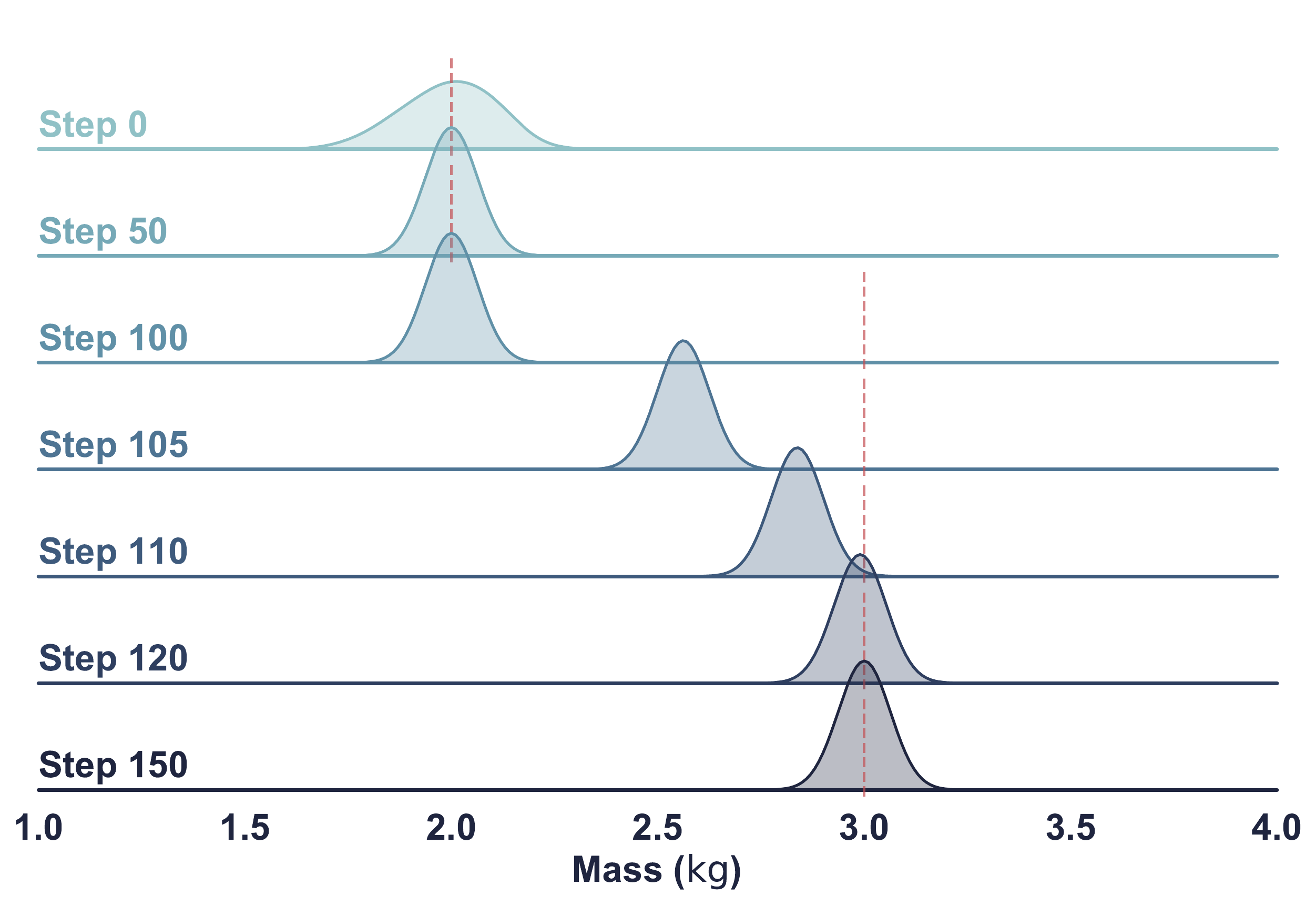}
    \caption{Ridge plot of mass distribution (in kg)}
    \label{fig:part2d_ridge_plot}
    \end{subfigure}
    \caption{\textbf{Point-mass navigation task}.
    The plots shows trajectories from the start position (red dot) towards the
    goal (red star).
    \textbf{Left}: Trajectories executed by \sv.
    Note that, as the mass of the robot changes, the model mismatch causes many
    of the episodes to crash (\textbf{x} markers).
    \textbf{Centre}: Trajectories executed by Dust-MPC.
    Depending on the state of the system when the mass change occurs, a few
    trajectories deviate from the centre path to avoid collisions.
    A few trajectories are truncated due to the fixed episode length.
    \textbf{Right}: Ridge plot of the distribution over mass along several steps
    of the simulation.
    The vertical dashed line denotes the true mass.
    Mass is initially set at 2 kg, and changed to 3 kg at step 100.}
    \label{fig:part2d_results}\vspace{-10pt}
\end{figure*}

In recent work~\cite{lambert_stein_2020, okada_variational_2019}, the MPC
problem has been formulated as a Bayesian inference task whose goal is to
estimate the posterior distribution over control parameters given the state and
observed costs.
To make such problem tractable in real-world applications,
variational inference has been used to approximate the optimal
distribution.
These approaches are better suited at handling the multi-modality of the
distribution over actions, but do not attempt to dynamically adapt to
changes in the environment.

Conversely, previous work has demonstrated that incorporating uncertainty
in the evaluation of SOC estimates can improve performance
\cite{barcelos_disco_2020}, particularly when this uncertainty is periodically
re-estimated ~\cite{possasOnlineBayesSimCombined2020}.
Although this method is more robust to model mismatch and help
address the sim-to-real gap, the strategy relies on applying moment-matching
techniques to propagate the uncertainty through the dynamical system which is 
approximated by a Gaussian distribution.
This diminishes the effectiveness of the method under settings where
multi-modality is prominent.

In this work, we aim to leverage recent developments in variational inference 
with MPC for decision-making under complex multi-modal uncertainty over actions
while simultaneously estimating the uncertainty over model parameters.
We propose a Stein variational stochastic gradient solution that models the
posterior distribution over actions and model parameters as a collection of
particles, representing an implicit variational distribution.
These particles are updated sequentially, online, in parallel, and can capture 
complex multi-modal distributions. 

Specifically, the main contributions of this paper are:
\begin{itemize}
    \item We propose a principled Bayesian solution of introducing uncertainty over model
    parameters in Stein variational MPC and empirically demonstrate how this can
    be leveraged to improve the control policy robustness;
    
    \item We introduce a novel method to extend the inference problem and
    simultaneously optimise the control policy while refining our knowledge of
    the environment as new observations are gathered.
    Crucially, by leveraging recent advancements in sequential Monte Carlo with
    kernel embedding, we perform online, sequential updates to the distribution
    over model parameters which scales to large datasets;
    
    \item By capturing the uncertainty on true dynamic systems in distributions
    over a parametric model, we are able to incorporate domain knowledge and
    physics principles while still allowing for a highly representative model.
    This simplifies the inference problem and drastically reduces the number of
    interactions with the environment to characterise the model.
\end{itemize}

We implement the algorithm on a real autonomous ground vehicle (AGV)
(\cref{fig:wombot}), illustrating the applicability of the method in real time.
Experiments show how the control and parameter inference are leveraged to
adapt the behaviour of the robot under varying conditions, such as changes in
mass.
We also present simulation results on an inverted pendulum and an 2D obstacle
grid, see \cref{fig:part2d_results}, demonstrating an effective adaptation to
dynamic changes in model parameters.

This paper is organised as follows. In Section~\ref{sec:related} we review
related work, contrasting the proposed method to the existing literature.
In Section~\ref{sec:background} we provide background on stochastic MPC and
Stein variational gradient descent as a foundation to the proposed method, which
is presented in Section~\ref{sec:method}.
In Section~\ref{sec:experiments} we present a number of real and simulated
experiments, followed by relevant conclusions in Section~\ref{sec:conclusion}.
\section{Related Work} \label{sec:related}
Sampling-based approaches for stochastic MPC have shown to be suitable for a range of control problems in robotics~\cite{williams_information-theoretic_2018,wagener2019online}. At each iteration, these methods perform an approximate evaluation by rolling-out a stochastic policy with modelled system dynamics over a finite-length horizon. The optimisation step proceeds to update the policy parameters in the direction that minimises the expected cost and a statistical distance to a reference policy or prior~\cite{wagener2019online}. This can equivalently be interpreted as a statistical inference procedure, where policy parameters are updated in order to match an optimal posterior distribution distribution~\cite{rawlik2012stochastic, williams_information-theoretic_2018, lambert_stein_2020}. This connection has motivated the use of common approximate inference procedures for SOC. Model Predictive Path Integral Control (MPPI)~\cite{williamsModelPredictivePath2017, williams_information-theoretic_2018} and the Cross Entropy Method (CEM)~\cite{cem}, for instance, use importance sampling to match a Gaussian proposal distribution to moments of the posterior. Variational inference (VI) approaches have also been examined for addressing control problems exhibiting highly non-Gaussian or multi-modal posterior distributions. This family of Bayesian inference methods extends the modelling capacity of control distribution to minimise a Kullback-Leibler divergence with the target distribution. Traditional VI methods such as Expectation-Maximisation have been examined for control problems~\cite{watson2020stochastic, okada_variational_2019}, where the model class of the probability distribution is assumed to be restricted to a parametric family (typically Gaussian Mixture Models).  More recently, the authors in ~\cite{lambert_stein_2020} proposed to adapt the Stein Variational Gradient Descent (SVGD)~\cite{liu_stein_2019} method for model predictive control.
Here, a distribution of control sequences is represented as a collection of particles. The resulting framework, Stein-Variational Model Predictive Control (\sv), adapts the particle distribution in an online fashion. The non-parametric representation makes the approach particularly suitable to systems exhibiting multi-modal posteriors. In the present work, we build on the approach in ~\cite{lambert_stein_2020} to develop an MPC framework which leverages particle-based variational inference to optimise controls \textit{and} explicitly address model uncertainty. Our method \textit{simultaneously} adapts a distribution over dynamics parameters, which we demonstrate improves robustness and performance on a real system.

Model predictive control is a reactive control scheme, and can accommodate modelling errors to a limited degree. However, its performance is largely affected by the accuracy of long-range predictions. Modelling errors can compound over the planning horizon, affecting the expected outcome of a given control action. This can be mitigated by accounting for model uncertainty, leading to better estimates of expected cost. This has been demonstrated to improve performance in stochastic optimal control methods and model-based reinforcement learning~\cite{pan2014probabilistic, deisenroth_pilco:_nodate}. Integrating uncertainty has typically been achieved by learning probabilistic dynamics models from collected state-transition data in an episodic setting, where the model is updated in between trajectory-length system executions~\cite{chua_deep_2018, okada_variational_2019, wabersich2020bayesian, ramos_bayessim:_2019}. A variety of modelling representations have been explored, including Gaussian processes~\cite{deisenroth_pilco:_nodate}, neural network ensembles~\cite{chua_deep_2018}, Bayesian regression, and meta-learning~\cite{harrison2018meta}. Alternatively, the authors in \cite{ramos_bayessim:_2019, possasOnlineBayesSimCombined2020} estimate posterior distributions of physical parameters for black-box simulators, given real-world observations.

Recent efforts have examined the online setting, where a learned probabilistic model is updated based on observations made \textit{during execution}~\cite{pan2016adaptive, abraham2020model, fan_deep_2020, fisac_general_2019}. The benefits of this paradigm are clear: incorporating new observations and adapting the dynamics \textit{in situ} will allow for better predictions, improved control, and recovery from sudden changes to the environment. However, real-time requirements dictate that model adaptation must be done quickly and efficiently, and accommodate the operational timescale of the controller. This typically comes at the cost of modelling accuracy, and limits the application of computationally-burdensome representations, such as neural networks and vanilla GPs. Previous work has included the use of sparse-spectrum GPs and efficient factorization to incrementally update the dynamics model ~\cite{pan2016adaptive, pan_prediction_2017}. In~\cite{harrison2018meta}, the authors use a meta-learning approach to train a network model offline, which is adapted to new observations using Bayesian linear regression operating on the last layer. However, these approaches are restricted to Gaussian predictive distributions, and may lack sufficient modelling power for predicting complex, multi-modal distributions.

Perhaps most closely related to our modeling approach is the work by Abraham et al.~\cite{abraham2020model}. The authors propose to track a distribution over simulation parameters using a sequential Monte Carlo method akin to a particle filter. The set of possible environments resulting from the parameter distribution is used by an MPPI controller to generate control samples. Each simulated trajectory rollout is then weighted according to the weight of the corresponding environment parameter. Although such an approach can model multi-modal posterior distributions, we should expect similar drawbacks to particle filters, which require clever re-sampling schemes to avoid mode collapse and particle depletion. Our method also leverages a particle-based representation of parameter distributions, but performs deterministic updates based on new information and is more sample efficient than MC sampling techniques.
\section{Background} \label{sec:background}
%%%%%%%%%%%%%%%%%%%%%%%%%%%%%%%%%%%%%%%%%%%%%%%%%%%%%%%%%%%%%%%%%%%%%%%%%%%%%%%%

%%%%%%%%%%%%%%%%%%%%%%%%%%%%%%%%%%%%%%%%%%%%%%%%%%%%%%%%%%%%%%%%%%%%%%%%%%%%%%%%
\subsection{Stochastic Model Predictive Control}
%%%%%%%%%%%%%%%%%%%%%%%%%%%%%%%%%%%%%%%%%%%%%%%%%%%%%%%%%%%%%%%%%%%%%%%%%%%%%%%%
We consider the problem of controlling a discrete-time nonlinear system
compactly represented as:
\begin{align} \label{eq:gen_problem}
    \state_{t+1} = \system(\state_{t}, \control_{t})
\end{align}
where $\system$ is the transition map, $\state_{t} \in \stateSpace$ denotes the
system state, $\control_{t} \in \controlSpace$ the control input, and 
$\stateSpace$ and $\controlSpace$ are respectively appropriate Euclidean spaces 
for the states and controls.
More specifically, we are interested in the problem where the real transition 
function $\system(\state_\tIdx, \control_\tIdx)$ is unknown and approximated by 
a transition function with \textit{parametric uncertainty}, such that:
\begin{equation}
    \system(\state_\tIdx, \control_\tIdx)
    \approx \hat{\system}(\state_\tIdx, \control_\tIdx, \simParams)
    =: \simFunction(\state_\tIdx, \control_\tIdx) 
\end{equation}
where $\simParams \in \simSpace$ are the simulator parameters with a prior 
probability distribution $\pDensity(\simParams)$ and the non-linear forward 
model $\hat{\system}(\state, \control, \simParams)$, is represented as 
$\simFunction$ for compactness.

Based on the steps in \cite{barcelos_disco_2020}, we can then define a fixed 
length control sequence $\controlSeq_{\tIdx} = \{\control_\tIdx\}_{\tIdx < 
\tIdx + \controlHorizon}$ over a fixed control horizon $\controlHorizon$, onto 
which we apply a receding horizon control (RHC) strategy.
Moreover, we define a mapping operator, $\seqOp$, from input sequences 
$\controlSeq_{\tIdx}$ to their resulting states by recursively applying 
$\simFunction$ given $\state_{\tIdx}$, 
\begin{equation} \label{eq:rollout_operator}
\begin{split}
    \stateSeq_{\tIdx} &= \seqOp(\controlSeq_{\tIdx}; \state_{\tIdx}) \\
    &= \left[\state_{\tIdx}, \simFunction(\state_{\tIdx}, \control_{\tIdx}),
    \simFunction(
    \simFunction(\state_{\tIdx}, \control_{\tIdx}), \control_{\tIdx + 1}),
    \ldots \right] \,, 
\end{split}
\end{equation}
noting that $\stateSeq_{\tIdx}$ and $\controlSeq_{\tIdx}$ define estimates of
future states and controls.
Finally, we can define a \emph{trajectory} as the joint sequence of states and 
actions, namely:
\begin{equation}
    \trajectory = \left(\stateSeq_{\tIdx}, \controlSeq_{\tIdx}\right) \,.
\end{equation}

In MPC, we want to minimise a task dependent cost functional 
generally described as:
\begin{equation} \label{eq:cost_function}
    \cost\left(\stateSeq_{\tIdx}, \controlSeq_{\tIdx}\right) =
    \termCost(\hat{\state}_{\tIdx + \controlHorizon})
    + \sum_{\hzIdx =  0}^{\controlHorizon - 1}
    \instCost(\hat{\state}_{\tIdx + \hzIdx}, \hat{\control}_{\tIdx + \hzIdx}) 
    \,,
\end{equation}
where $\hat{\state}$ and $\hat{\control}$ are the estimated states and controls,
and $\instCost(\cdot)$ and  $\termCost(\cdot)$ are respectively arbitrary 
instantaneous and terminal cost functions.
To do so, our goal is to find an optimal policy $\policy(\state_{\tIdx})$ to
generate the control sequence $\controlSeq_{\tIdx}$ at each time-step.
As in \cite{wagener2019online}, we define such feedback policy as a 
parameterised probability distribution $\policy_{\polParams_{\tIdx}} = 
\pDensity(\control_{\tIdx} \given \state_{\tIdx}; \polParams_{\tIdx})$ from 
which the actions at time $\tIdx$ are sampled, \ie $\controlSeq_{\tIdx} \sim
\policy_{\polParams_{\tIdx}}$.
The control problem can then be defined as:
\begin{equation}
    \min_{\polParams_{\tIdx} \in \polSpace}
    \hat{J}\left(\policy_{\polParams_{\tIdx}}; \state_{\tIdx}\right)\,,
\end{equation}
where $\hat{J}$ is an estimator for a statistic of interest, typically
$\expectation_{\policy_{\polParams_{\tIdx}}, \simFunction}
\left[\cost\left(\stateSeq_{\tIdx}, \controlSeq_{\tIdx}\right)\right]$.
Once $\policy_{\polParams_{\tIdx}}$ is determined, we can use it to sample 
$\controlSeq_{\tIdx}$, extract the first control $\control_{\tIdx}$ and apply it
to the system.

One of the main advantages of stochastic MPC is that $\hat{J}$ can be computed
even when the cost function is non-differentiable \wrt to the policy parameters
by using importance sampling~\cite{williamsModelPredictivePath2017}.

%%%%%%%%%%%%%%%%%%%%%%%%%%%%%%%%%%%%%%%%%%%%%%%%%%%%%%%%%%%%%%%%%%%%%%%%%%%%%%%%
\subsection{Stein Variational Gradient Descent}
\label{sec:svgd}
%%%%%%%%%%%%%%%%%%%%%%%%%%%%%%%%%%%%%%%%%%%%%%%%%%%%%%%%%%%%%%%%%%%%%%%%%%%%%%%%

Variational inference poses posterior estimation as an optimisation task where
a candidate distribution $\qDensity^{*}(\anyvector)$ within a distribution
family $\qClass$ is chosen to best approximate the target distribution
$p(\anyvector)$.
This is typically obtained by minimising the Kullback-Leibler (KL) divergence:
\begin{equation} \label{eq:vi_obj}
    \qDensity^{*}(\anyvector) = \argmin_{\qDensity \in \qClass} \
    \kl {\qDensity(\anyvector)} {p(\anyvector)} \,.
\end{equation}

The solution also maximises the Evidence Lower Bound (ELBO), as expressed by the
following objective:
\begin{equation}
    \qDensity^{*}(\anyvector) = \argmax_{\qDensity \in \qClass}
    \expectation_{\qDensity} [\log p(\anyvector)] 
    - \kl {\qDensity(\anyvector)} {p(\anyvector)}
\end{equation}

In order to circumvent the challenge of determining an appropriate $\qClass$,
while also addressing \cref{eq:vi_obj}, we develop an algorithm based on Stein
variational gradient descent (SVGD) for Bayesian
inference~\cite{liu_stein_2019}.
The non-parametric nature of SVGD is advantageous as it removes the need for
assumptions on restricted parametric families for $\qDensity(\anyvector)$.
This approach approximates a posterior $p(\anyvector)$ with a set of particles 
$\{\anyvector^\partIdx\}_{\partIdx}^{\partSize}$, $\anyvector \in \R^\partDim$.
These particles are iteratively updated according to: 
\begin{equation} \label{eq:stein_update}
\anyvector^{\partIdx} \leftarrow \anyvector^{\partIdx} + 
     \stepSize \scoreFunc^{*}(\anyvector^{\partIdx})\,, 
\end{equation} 
given a step size $\stepSize$.
The function $\phi(\cdot)$ is known as score function and defines the velocity field that maximally decreases the KL-divergence.
\begin{equation}
    \scoreFunc^{*} = 
    \underset {\scoreFunc \in \steinRKHS} {\argmax}
    \left\{
    -\nabla_{\stepSize} 
    \kl{\qDensity_{[\stepSize \phi]}}{p(\anyvector)},
    \text{~\st~} \|\scoreFunc\|_{\steinRKHS} \leq 1
    \right\}
\end{equation}
where $\steinRKHS$ is a Reproducing Kernel Hilbert Space (RKHS) induced by the kernel function used and $\qDensity_{[\stepSize \phi]}$ indicates the particle distribution resulting from taking an update step as in \cref{eq:stein_update}.
In \cite{liu_stein_2019} this has been shown to yield a closed-form solution which can be interpreted as a functional gradient in $\steinRKHS$ and approximated with the set of particles:
\begin{equation} \label{eq:stein_approx_score_func}
    {\scoreFunc}^* (\anyvector) = \frac{1}{\partSize} \sum_{\secIdx = 1}^{\partSize}
    \left[
    k(\anyvector^{\secIdx}, \anyvector)
    \nabla_{\anyvector^{\secIdx}}
    \log p(\anyvector^{\secIdx})
    + \nabla_{\anyvector^{\secIdx}}
    k(\anyvector^{\secIdx}, \anyvector)
    \right]
\end{equation}

\section{Method} \label{sec:method}
%%%%%%%%%%%%%%%%%%%%%%%%%%%%%%%%%%%%%%%%%%%%%%%%%%%%%%%%%%%%%%%%%%%%%%%%%%%%%%%%
In this section, we present our approach for joint inference over control and 
model parameters for MPC. We call this method Dual Stein Variational Inference
MPC, or \dust for conciseness.
We begin by formulating optimal control as an inference problem and address how
to optimise policies in \cref{sec:policy_inference}. 
Later, in \cref{sec:dyn_inference}, we extend the inference to also include the
system dynamics.
A complete overview of the method is described in algorithm in
\cref{app:algo}. 

%%%%%%%%%%%%%%%%%%%%%%%%%%%%%%%%%%%%%%%%%%%%%%%%%%%%%%%%%%%%%%%%%%%%%%%%%%%%%%%%
\subsection{MPC as Bayesian Inference}
\label{sec:bayesian_mpc}
%%%%%%%%%%%%%%%%%%%%%%%%%%%%%%%%%%%%%%%%%%%%%%%%%%%%%%%%%%%%%%%%%%%%%%%%%%%%%%%%

MPC can be framed as a Bayesian inference problem where we estimate
the posterior distribution of policies, parameterised by 
$\polParams_{\tIdx}$, given an optimality criterion.
% We start by defining a trajectory as a sequence of states and actions of length
% $H+1$, namely: $\vec\tau = \{\state_{t+h},  \control_{t+h}\}_{h=0}^\controlHorizon$.
Let $\optimality : \trajectorySpace \to \{0, 1\}$ be an \emph{optimality}
indicator for a trajectory $\trajectory \in \trajectorySpace$ such that
$\optimality[\trajectory] = 1$ indicates that the trajectory is optimal.
Now we can apply Bayes' Rule and frame our problem as estimating:
\begin{equation} \label{eq:bayes_mpc}
    p(\polParams_{\tIdx} \given \optimality)
    \propto p(\optimality \given \polParams_{\tIdx}) p(\polParams_{\tIdx})
    = p(\optimality, \polParams_{\tIdx}).
\end{equation}

% In essence, optimality is a subjective term which might be difficult to 
% define.
% However, we may agree that in the MPC formulation the optimality of a 
% trajectory is related to the task-dependent cost functional defined in 
% \cref{sec:background}.
A reasonable to way to quantify $\optimality[\trajectory]$ is to model it as a
Bernoulli random variable conditioned on the trajectory $\trajectory$, allowing
us to define the \textit{likelihood} $p(\optimality \given \polParams_{\tIdx})$
as:
\begin{equation} \label{eq:svmpc_likelihood}
\begin{split}
    p(\optimality \given \polParams_{\tIdx})
    % &:= P(\trajectorySpace^* \given \polParams_{\tIdx}) \\
    &= \expectation_{\trajectory \sim p(\trajectory \given \polParams_{\tIdx})}
    \left[p(\optimality[\trajectory] = 1 \given \trajectory)\right] \\
    &\approx \frac{1}{n} \sum_{i=1}^n \exp(-\alpha \cost[\trajectory_i])\,,
\end{split}
\end{equation}
where $\trajectory_i \overset{\iid}{\sim} p(\trajectory \given 
\polParams_{\tIdx})$, $i  \in \{1, 2, \dots, n\}$,
$p(\optimality[\trajectory] = 1 \given \trajectory) := \exp(-\alpha 
\cost[\trajectory])$, and we overload $p(\optimality[\trajectory] = 1 \given
\polParams_{\tIdx})$ to simplify the notation.
% Taking $p(\optimality \given \polParams_{\tIdx})$ as a \emph{likelihood} 
% function
Now, assuming a prior $p(\polParams_{\tIdx})$ for $\polParams_{\tIdx}$,
the posterior over $\polParams_{\tIdx}$ is given by:
\begin{equation}
\begin{split}
    p(\polParams_{\tIdx} \given \optimality)
    &\propto p(\optimality \given \polParams_{\tIdx}) p(\polParams_{\tIdx})= \int_\trajectorySpace p(\optimality \given \trajectory) 
    p(\trajectory \given \polParams_{\tIdx}) p(\polParams_{\tIdx}) \diff\trajectory \,.
\end{split}
\end{equation}
This posterior corresponds to the probability (density) of a given parameter
setting  $\polParams_{\tIdx}$ conditioned on the hypothesis that (implicitly 
observed) trajectories generated by $\polParams_{\tIdx}$ are \emph{optimal}.
Alternatively, one may say that $p(\polParams_{\tIdx} \given \optimality)$ tells
us the probability of $\polParams_{\tIdx}$ being the generator of the optimal 
set $\trajectorySpace^*$.
Lastly, note that the trajectories conditional $p(\trajectory \given 
\polParams_{\tIdx})$ factorises as:
\begin{equation}
    p(\trajectory \given \polParams_{\tIdx})
    = \prod_{\hzIdx=0}^{\controlHorizon - 1}
    \simDensity(
    \state_{\tIdx + \hzIdx + 1}
    \given \state_{\tIdx + \hzIdx}, \control_{\tIdx + \hzIdx}
    )
    \policy_{\polParams_{\tIdx}}(\state_{\tIdx + \hzIdx})\,. 
\end{equation}

%%%%%%%%%%%%%%%%%%%%%%%%%%%%%%%%%%%%%%%%%%%%%%%%%%%%%%%%%%%%%%%%%%%%%%%%%%%%%%%%
\subsection{Joint Inference for Policy and Dynamics with Stein MPC}
%%%%%%%%%%%%%%%%%%%%%%%%%%%%%%%%%%%%%%%%%%%%%%%%%%%%%%%%%%%%%%%%%%%%%%%%%%%%%%%%

In this section, we generalise the framework presented in
\cite{lambert_stein_2020} to simultaneously refine our knowledge of the
dynamical system, parameterised by $\simParams$, while estimating optimal policy
parameters $\polParams_{\tIdx}$.
Before we proceed, however, it is important to notice that the optimality
measure defined in \cref{sec:bayesian_mpc} stems from \textit{simulated}
rollouts sampled according to \cref{eq:svmpc_likelihood}.
Hence, these trajectories are not actual observations of the agent's environment
and are not suitable for inferring the parameters of the system dynamics.
In other words, to perform inference over the parameters $\simParams$
we need to collect \textit{real} observations from the environment.

With that in mind, the problem statement in \cref{eq:bayes_mpc} can be
rewritten as inferring: 
\begin{equation} \label{eq:dust_problem}
    \pDensity(\polParams_{\tIdx}, \simParams \given \optimality, \dataset)
    = \pDensity(\polParams_{\tIdx} \given \optimality, \simParams)
    \pDensity(\simParams \given \dataset)\,,
\end{equation}
where $\dataset := \{(\state_{\tIdx}^r, \control_{\tIdx - 1}^r,
\state_{\tIdx  - 1}^r)\}_{\tIdx = 1}^{\dataSize}$ represents the dataset of
collected environment observations.
Note that the policy parameters $\polParams$ are independent from the system
observations, and therefore the conditioning on $\dataset$ has been dropped.
Similarly, the distribution over dynamics parameters is independent from
$\optimality$ and again we omit the conditioning.

The reader might wonder why we have factorised \cref{eq:dust_problem} instead 
of defining a new random variable $\Theta = \{\polParams, \simParams\}$, and
following the steps outlined in~\cite{lambert_stein_2020} to solve the joint 
inference problem.
By jointly solving the inference problem the partial derivative of $\simParams$
would be comprised of terms involving both factors on the right-hand side (RHS)
of \cref{eq:dust_problem}, biasing the estimation of $\simParams$ to regions
of lower control cost (see \cref{app:joint_bias} for further discussion).
Furthermore, the two inference problems are naturally disjoint. While the 
policy inference relies on simulated future rollouts, the dynamics inference is
based solely on previously observed data.

By factorising the problem we can perform each inference step
separately and adjust the computational effort and hyper-parameters based on the
idiosyncrasies of the problem at hand.
This formulation is particularly amenable for stochastic MPC, as seen 
in \cref{sec:background}, since the policy $\policy_{\polParams_{\tIdx}}$ is 
independent of the approximated transition function $\simFunction$, \ie 
$\polParams \indep \simParams$.

%%%%%%%%%%%%%%%%%%%%%%%%%%%%%%%%%%%%%%%%%%%%%%%%%%%%%%%%%%%%%%%%%%%%%%%%%%%%%%%%
\subsection{Policy Inference for Bayesian MPC}
\label{sec:policy_inference}
%%%%%%%%%%%%%%%%%%%%%%%%%%%%%%%%%%%%%%%%%%%%%%%%%%%%%%%%%%%%%%%%%%%%%%%%%%%%%%%%

Having formulated the inference problem as in \cref{eq:dust_problem}, we can
proceed by solving each of the factors separately.
We shall start with optimising $\policy_{\polParams_{\tIdx}}$ according to
$\pDensity(\polParams_{\tIdx} \given \optimality, \simParams)$.
It is evident that we need to consider the dynamics parameters when optimising
the policy, but at this stage we may simply assume the inference over 
$\simParams$ has been solved and leverage our knowledge of 
$\pDensity(\simParams|\dataset[t])$.
More concretely, let us rewrite the factor on the RHS of
\cref{eq:dust_problem} by marginalising over $\simParams$ so that:
\begin{equation} \label{eq:dust_pol_inf_dataset}
    \pDensity(\polParams_{\tIdx} \given \optimality, \dataset)
    \propto \int_\simSpace \likelihood_\policy(\polParams_{\tIdx}) 
    \pDensity(\polParams_{\tIdx}) \pDensity(\simParams \given \dataset) 
    \diff\simParams \,,
\end{equation}
where, as in \cref{eq:svmpc_likelihood}, the likelihood 
$\likelihood_\policy(\polParams_{\tIdx})$ is defined as:
\begin{equation} \label{eq:dust_pol_lik}
\begin{split}
    \likelihood_\policy(\polParams_{\tIdx})
    &= \pDensity(\optimality \given \polParams_{\tIdx}, \simParams)
    := P(\trajectorySpace^* \given \polParams_{\tIdx}, \simParams) \\
    &= \int_\trajectorySpace \pDensity(\optimality \given \trajectory)
    \pDensity(\trajectory \given \polParams_{\tIdx}, \simParams)
    \diff\trajectory \\
    &\approx \frac{1}{\trajectorySamples} \sum_{\primIdx = 1}^{\trajectorySamples}
    \exp(-\alpha \cost[\trajectory_\primIdx]) \,,
\end{split}
\end{equation}
with $\trajectory_\primIdx \overset{\iid}{\sim} \pDensity(\trajectory \given
\polParams_{\tIdx}, \simParams)$, $\primIdx \in \{1, 2, \dots, \trajectorySamples\}$,
and $\pDensity(\simParams \given \dataset)$ defines the inference problem of
updating the posterior distribution of the dynamics parameters given all
observations gathered from the environment, which we shall discuss in the
next section.
Careful consideration of \cref{eq:dust_pol_lik} tells us that unlike the case
of a deterministic transition function or even maximum likelihood point 
estimation of $\simParams$, the optimality of a given trajectory now depends on
its \textit{expected} cost over the distribution $\pDensity(\simParams)$.

Hence, given a prior $\pDensity(\polParams_{\tIdx})$ and $\pDensity(\simParams 
\given \dataset)$, we can generate samples from the likelihood in  
\cref{eq:dust_pol_lik} and use an stochastic gradient method as in 
\cref{sec:svgd} to infer the posterior distribution over $\polParams_{\tIdx}$.
% Inspired by the work in~\cite{pulido_sequential_2019}, we can combine ideas of
% optimal transport and particle flows to sequentially update the posterior
% distribution in \cref{eq:dust_dyn_inf}.
Following the steps in~\cite{lambert_stein_2020}, we approximate the prior
$\pDensity(\polParams_{\tIdx})$ by a set of particles 
$\qDensity(\polParams_{\tIdx}) = \{\polParams_{\tIdx}\}_{\primIdx = 1}^\polSize$ 
and take sequential SVGD updates:
\begin{equation} \label{eq:dust_pol_step}
    \polParams_{\tIdx}^\primIdx \gets \polParams_{\tIdx}^\primIdx 
    + \stepSize \scoreFunc^*(\polParams_{\tIdx}^\primIdx) \,,
\end{equation}
to derive the posterior distribution.
Where, again, $\scoreFunc^*$ is computed as in \cref{eq:stein_approx_score_func}
for each intermediate step and $\stepSize$ is a predetermined step size.
One ingredient missing to compute the score function is the gradient of the 
log-posterior of  $\pDensity(\polParams_{\tIdx} \given \optimality, 
\simParams)$, which can be factorised into:
\begin{equation} \label{eq:dust_pol_grad}
    \nabla_{\polParams_{\tIdx}^\primIdx} 
    \log \pDensity(\polParams_{\tIdx}^\primIdx \given \optimality, \simParams) 
    = \nabla_{\polParams_{\tIdx}^\primIdx}
    \log\likelihood(\polParams_{\tIdx}^\primIdx)
    + \nabla_{\polParams_{\tIdx}^\primIdx} 
    \log\qDensity(\polParams_{\tIdx}^\primIdx) \,.
\end{equation}
In practice we typically assume that the $\cost[\cdot]$ functional used as
surrogate for optimality is not differentiable \wrt $\polParams_{\tIdx}$, but
the gradient of the log-likelihood can be usually approximated via Monte-Carlo 
sampling.

Most notably, however, is the fact that unlike the original formulation in SVGD,
the policy distribution we are trying to infer is time-varying and depends on 
the actual state of the system.
The measure $P(\trajectorySpace^* \given \polParams_{\tIdx}; 
\state_\tIdx)$ depends on the current state, as that is the initial condition 
for all trajectories evaluated in the cost functional $\cost[\trajectory]$.

Theoretically, one could choose an uninformative prior with broad support
over the $\controlSpace$ at each time-step and employ standard SVGD to compute
the posterior policy.
However, due to the sequential nature of the MPC algorithm, it is likely that
the prior policy $\qDensity(\polParams_{\tIdx - 1})$ is close to a region of low
cost and hence is a good candidate to bootstrap the posterior inference of the
subsequent step.
This procedure is akin to the prediction step commonly found in Sequential
Bayesian Estimation~\cite{doucetSequentialMonteCarlo2001}.
More concretely,
\begin{equation} \label{eq:dust_pol_shift}
    \qDensity(\polParams_\tIdx)
    = \int \pDensity(\polParams_\tIdx \given \polParams_{\tIdx - 1})
    \qDensity(\polParams_{\tIdx - 1}) \diff\polParams_{\tIdx - 1} \,,
\end{equation}
where $\pDensity(\polParams_{\tIdx} \given \polParams_{\tIdx - 1})$ can be an
arbitrary transitional probability distribution.
More commonly, however, is to use a probabilistic version of the shift
operator as defined in~\cite{wagener2019online}.
For a brief discussion on action selection, please refer to
\cref{app:action_selection}.
For more details on this section in general the reader is encouraged to refer
to~\cite[Sec. 5.3]{lambert_stein_2020}.

\subsection{Real-time Dynamics Inference}
\label{sec:dyn_inference}
%%%%%%%%%%%%%%%%%%%%%%%%%%%%%%%%%%%%%%%%%%%%%%%%%%%%%%%%%%%%%%%%%%%%%%%%%%%%%%%%

We now focus on the problem of updating the posterior over the simulator
parameters.
Note that, due to the independence of each inference problem, the frequency
in which we update $\pDensity(\simParams)$ can be different from the policy
update.
% In fact, as discussed in \cref{sec:related}, many previous works rely on this
% to refine the parametric transition function in an \textit{episodic} setting
% \cite{okada_variational_2019, possasOnlineBayesSimCombined2020, ramos_bayessim:_2019}.
% Then, recursively, the new parameter distribution can be used during a new
% episode of data collection.

In contrast, we are interested in the case where $\pDensity(\simParams \given 
\dataset)$ can be updated in \textit{real-time} adjusting to changes in the 
environment.
For that end, we need a more efficient way of updating our posterior 
distribution.
The inference problem at a given time-step can then be written as:
\begin{equation} \label{eq:dust_dyn_inf_dataset}
\begin{split}
    \pDensity(\simParams \given \dataset)
    &\propto \pDensity(\datapoint \given \simParams, \dataset[\tIdx - 1])
    \pDensity(\simParams \given \dataset[\tIdx - 1]) \,.
    % &\propto \pDensity(\datapoint \given \simParams, \datapoint[\tIdx - 1])
    % \pDensity(\simParams \given \dataset[\tIdx - 1]) \,,
\end{split}
\end{equation}
Note that in this formulation, $\simParams$ is considered time-invariant.
This based on the implicit assumption that the frequency in which we gather new 
observations is significantly larger than the covariate shift to which 
$p(\simParams \given \dataset)$ is subject to as we traverse the environment.
In \cref{app:dyn_evolution}, we discuss the implications of changes in the
the latent parameter over time.

In general, we do not have access to direct measurements of $\simParams$, only
to the system state.
Therefore, in order to perform inference over the dynamics parameters, we rely 
on a generative model, \ie the simulator $\simFunction$, to generate samples in 
the state space $\stateSpace$ for different values of $\simParams$.
However, unlike in the policy inference step, for the dynamics parameter
estimation we are not computing deterministic simulated rollouts, but rather 
trying to find the explanatory parameter for each observed transition in the 
environment.
Namely, we have:
\begin{equation}
    \state^{r}_{\tIdx} = \system(\state^{r}_{\tIdx - 1}, \control_{\tIdx - 1})
    = \simFunction(\state^{r}_{\tIdx - 1}, \control_{\tIdx - 1}) + \noise_\tIdx
    \,,
\end{equation}
where $\state^{r}_{\tIdx}$ denotes the true system state and $\noise_\tIdx$ is a
time-dependent random variable closely related to the \textit{reality gap} in
the sim-to-real literature \cite{valassakisCrossingGapDeep2020} and incorporates
all the complexities of the real system not captured in simulation, such as
model mismatch, unmodelled dynamics, etc. 
As result, the distribution of $\noise_\tIdx$ is unknown, correlated over time
and hard to estimate.

In practice, for the feasibility of the inference problem, we make the standard
assumption that the noise is distributed according to a time-invariant normal
distribution $\noise_\tIdx \sim \normal(0, \mpfLikCov)$, with an empirically
chosen covariance matrix.
More concretely, this allows us to define the likelihood term in 
\cref{eq:dust_dyn_inf_dataset} as:
\begin{equation} \label{eq:dust_dyn_lik}
\begin{split}
    \likelihood(\simParams \given \dataset) 
    &:= \pDensity(\datapoint \given \simParams, \dataset[\tIdx - 1]) \\
    &= \pDensity(\state^{r}_{\tIdx} %= \state_{\tIdx}
    \given \simParams, \dataset[\tIdx - 1]) \\
    &= \normal(\state^{r}_{\tIdx}
    ; \simFunction(\state^{r}_{\tIdx - 1}, \control_{\tIdx - 1}), \mpfLikCov) \,,
\end{split}
\end{equation}
where we leverage the symmetry of the Gaussian distribution to centre the
uncertainty around $\state^{r}_{\tIdx}$.
It follows that, because the state transition is Markovian, the likelihood of
$\likelihood(\simParams | \dataset)$ depends only on the current observation 
tuple given by $\datapoint$, and we can drop the conditioning on previously
observed data.
In other words, we now have a way to quantify how likely is a given
realisation of $\simParams$ based on the data we have collected from the
environment.
Furthermore, let us define a single observation $\datapoint = (\state_{\tIdx}^r,
\control_{\tIdx - 1}^r, \state_{\tIdx - 1}^r)$ as the tuple of last applied
control action and observed state transition.
Inferring exclusively over the current state is useful whenever frequent 
observations are received and prevents us from having to store information over 
the entire observation dataset.
This approach is also followed by ensemble Kalman filters and particle flow 
filters~\cite{leeuwenParticleFiltersHigh2019}.

Equipped with \cref{eq:dust_dyn_lik} and assuming that an initial prior 
$p(\simParams)$ is available, we can proceed as discussed in 
\cref{sec:svgd} by approximating each prior at time $\tIdx$ with a set of particles $\{\simParams^\primIdx\}_{\primIdx = 1}^\simSize$ following
$\qDensity(\simParams \given \dataset[\tIdx - 1])$, so that our posterior over 
$\simParams$ at a given time $\tIdx$ can then be  rewritten as:
\begin{equation}
    \pDensity(\simParams \given \dataset)
    \approx \qDensity(\simParams \given \dataset)
    \propto \likelihood(\simParams \given \datapoint)
    \qDensity(\simParams \given \dataset[\tIdx - 1]) \,,
\end{equation}
and we can make recursive updates to $\qDensity(\simParams \given \dataset)$
by employing it as the prior distribution for the following step.
Namely, we can iteratively update $\qDensity(\simParams \given \dataset)$ a
number of steps $\mpfSteps$ by applying the update rule with a  functional gradient computed as in
\cref{eq:stein_approx_score_func}, where:
\begin{equation} \label{eq:dust_dyn_grad}
    \nabla_{\simParams}
    \log\pDensity_\tIdx(\simParams \given \dataset)
    \approx \nabla_{\simParams}
    \log\pDensity(\datapoint \given \simParams)
    + \nabla_{\simParams} \log\qDensity_\tIdx(\simParams\given\dataset) \,.
\end{equation}
An element needed to evaluate \cref{eq:dust_dyn_grad} is an expression for the
gradient of the posterior density.
An issue in sequential Bayesian inference is that there is no exact expression 
for the posterior density~\cite{pulido_sequential_2019}.
Namely, we know the likelihood function, but the prior density is only 
represented by a set of particles, not the density itself.

One could forge an empirical distribution $\qDensity(\simParams \given 
\dataset) = \frac{1}{\simSize}\sum_{\primIdx = 1}^\simSize \dirac{\simParams^\primIdx}$ by 
assigning Dirac functions at each particle location, but we would still be 
unable to differentiate the posterior.
In practice, we need to apply an efficient density estimation method, as we 
need to compute the density at each optimisation step.
We choose to approximate the posterior density with an equal-weight Gaussian 
Mixture Model (GMM) with a fixed diagonal covariance matrix:
\begin{equation} \label{eq:dust_dyn_posterior}
    \qDensity(\simParams \given \dataset) = \frac{1}{\simSize}
    \sum_{\primIdx = 1}^\simSize \normal(\simParams;
    \simParams^\primIdx, \mpfPriorCov) \,,
\end{equation}
where the covariance matrix $\mpfPriorCov$ can be predetermined or computed
from data.
One option, for example, is to use a Kernel Density bandwidth
estimation heuristic, such as Improved Sheather 
Jones~\cite{botevKernelDensityEstimation2010a},
Silverman's~\cite{silvermanDensityEstimationStatistics1986} or 
Scott's~\cite{scottMultivariateDensityEstimation1992} rule, to determine the
standard deviation $\sigma$ and set $\mpfPriorCov = \sigma^2 \eye$.

One final consideration is that of the support for the prior distribution.
Given the discussion above, it is important that the density approximation of
$\qDensity(\simParams \given \datapoint)$ offers support on regions of interest
in the parameter space.
In our formulation, that can be controlled by adjusting $\mpfPriorCov$.
Additionally, precautions have to be taken to make sure the parameter space is
specified correctly, such as using log-transformations for strictly positive
parameters for instance.
\section{Experiments}
\label{sec:experiments}
%%%%%%%%%%%%%%%%%%%%%%%%%%%%%%%%%%%%%%%%%%%%%%%%%%%%%%%%%%%%%%%%%%%%%%%%%%%%%%%%
In the following section we present experiments, both in simulation and with a 
physical autonomous ground vehicle (AGV), to demonstrate the correctness and
applicability of our method.

\subsection{Inverted pendulum with uncertain parameters}
\label{sec:exp_pendulum}
%%%%%%%%%%%%%%%%%%%%%%%%%%%%%%%%%%%%%%%%%%%%%%%%%%%%%%%%%%%%%%%%%%%%%%%%%%%%%%%%
\begin{figure}
    % get figure heights
    \sbox\twosubbox{
      \resizebox{\dimexpr\linewidth-1em}{!}{
        \includegraphics[height=3cm]{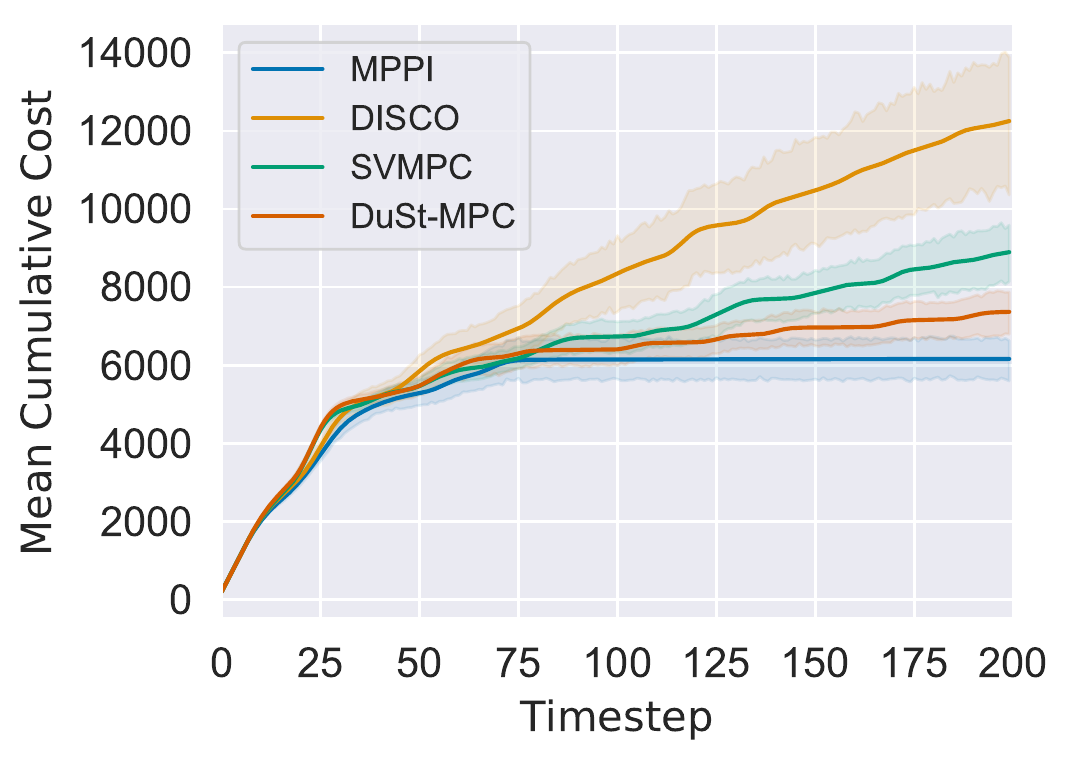}
        \includegraphics[height=3cm]{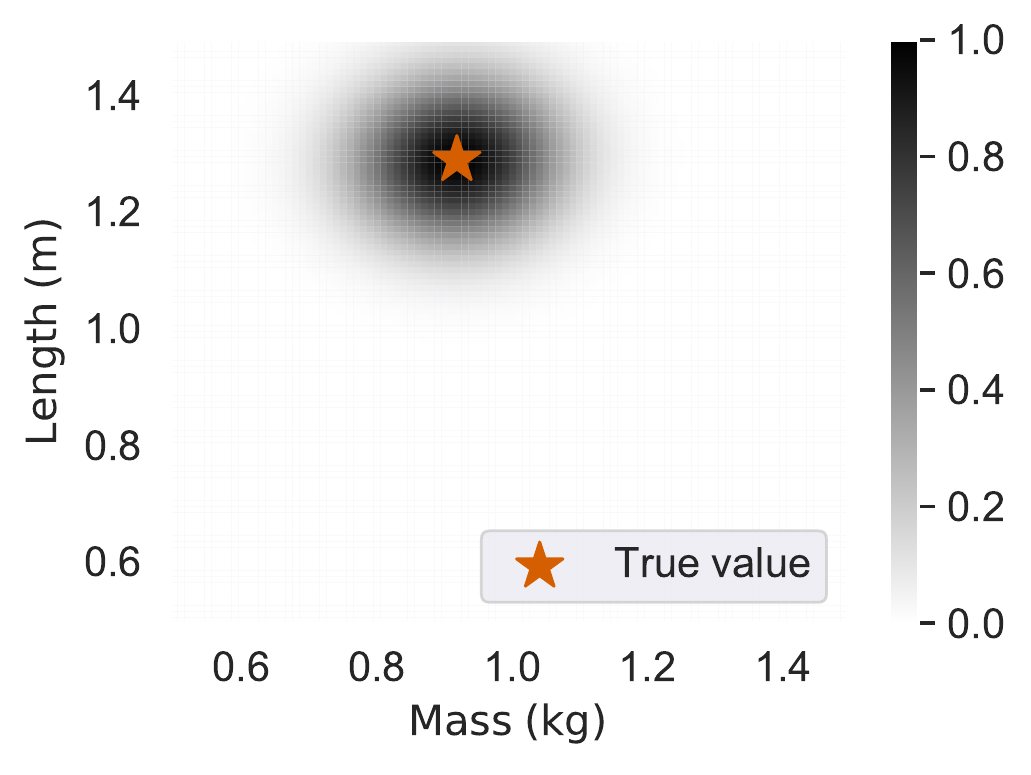}
      }
    }
    \setlength{\twosubht}{\ht\twosubbox}
    \centering
    \subcaptionbox{Pendulum costs\label{fig:pendulum_costs}}{
    \includegraphics[height=\twosubht]{fig/pendulum_costs_ci50.pdf}
    }
    \subcaptionbox{Pendulum posterior\label{fig:pendulum_posterior}}{
    \includegraphics[height=\twosubht]{fig/pendulum_posterior.pdf}
    }
    \caption{\textbf{Inverted pendulum}.
    (a) The image shows the mean cumulative cost over 10 episodes.
    The shaded region represents the 50\% confidence interval.
    The high variance is expected since each scenario has parameters sampled
    from a uniform distribution.
    (b) Plot of the posterior distribution over the pendulum
    pole-mass at the final step of one of the episodes.
    The true latent value is shown by the red star marker.}
    \label{fig:pendulum_results}\vspace{-10pt}
\end{figure}

We first investigate the performance of \dust in the classic inverted pendulum
control problem.
As usual, the pendulum is composed of a rigid pole-mass system controlled at one
end by a 1-degree-of-freedom torque actuator.
The task is to balance the point-mass upright, which, as the controller is
typically under-actuated, requires a controlled swing motion to overcome
gravity.
Contrary to the typical case, however, in our experiments the mass and length
of the pole-mass are unknown and equally likely within a range of 
\SIrange{0.5}{1.5}{\kilogram} and \SIrange{0.5}{1.5}{\meter}, respectively.

At each episode, a set of latent model parameters is sampled and used in the
simulated environment.
Each method is then deployed utilising this same parameter set.
MPPI is used as a baseline and has \textit{perfect knowledge} of the latent
parameters.
This provides a measure of the task difficulty and achievable results.
As discussed in \cref{sec:related}, we compare against \disco and
\sv as additional baselines.
We argue that, although these methods perform no online update of their
knowledge of the world, they offer a good underpinning for comparison since
the former tries to leverage the model uncertainty to create more
robust policies, whereas the latter shares the same variational inference
principles as our method.
\disco is implemented in its unscented transform variant applied to the
uninformative prior used to generate the random environments.
\sv uses the mean values for mass and length as point-estimates for its fixed
parametric model.
For more details on the hyper-parameters used, refer to \cref{app:experiments}.

\Cref{fig:pendulum_costs} presents the average cumulative costs over 10 
episodes.
Although the results show great variance, due to the randomised environment, it
is clear that \dust outperforms both baselines.
Careful consideration will show that the improvement is more noticeable as the
episode progresses, which is expected as the posterior distribution over the
model parameters being used by \dust gets more informative.
The final distribution over mass and length for one of the episodes is shown in
\cref{fig:pendulum_posterior}.
Finally, a summary of the experimental results is presented in
\cref{tab:sim_results}.

\begin{figure}[!ht]
    \sbox\twosubbox{
      \resizebox{\dimexpr\linewidth-1em}{!}{
        \includegraphics[height=3cm]{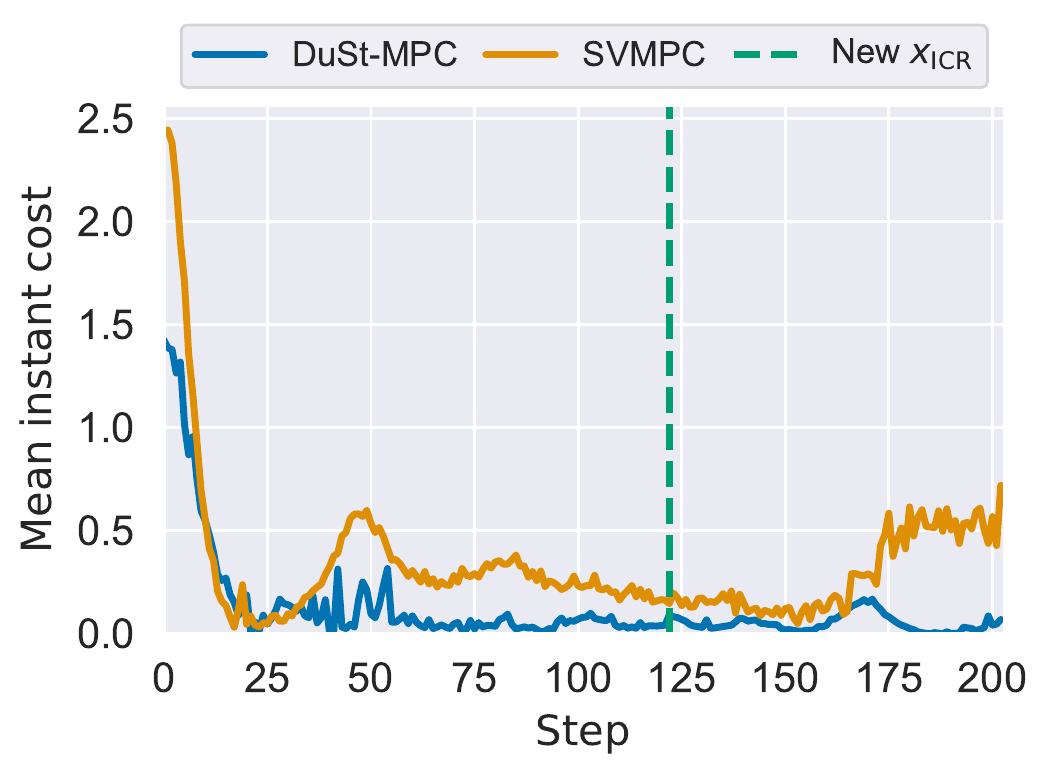}
        \includegraphics[height=3cm]{fig/wombot_raw_costs.pdf}
      }
    }
    \setlength{\twosubht}{\ht\twosubbox}
    \centering
    \subcaptionbox{Raw costs\label{fig:wombot_costs}}{
    \includegraphics[height=\twosubht]{fig/wombot_raw_costs.pdf}
    }
    \subcaptionbox{Trajectories\label{fig:wombot_trajectories}}{
    \includegraphics[height=\twosubht]{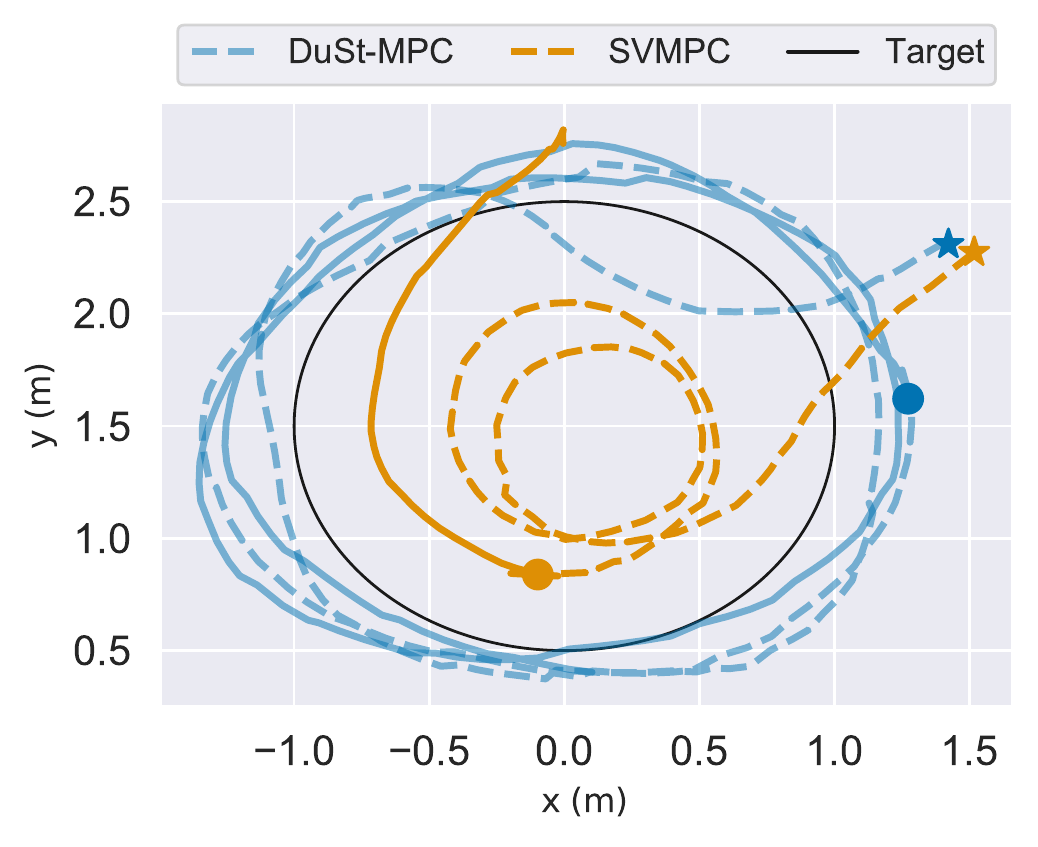}
    }
    \caption{\textbf{AGV trajectory tracking}.
    (a) Raw cost over time.
    Amount of steps before and after the change of mass are normalised for
    proper comparison.
    (b) Trajectories executed by each method.
    Line style changes when mass changes.
    Markers denote initial and change of mass position.
    }
    \label{fig:wombot_results}
\end{figure}

\begin{table}[t]
    \centering
    \begin{tabular}{lcccc} \toprule
              & \multicolumn{2}{c}{Point-mass} & \multicolumn{2}{c}{Pendulum}
              \\\cmidrule(lr){2-3}               \cmidrule(lr){4-5}
              & Cost $(\mu \pm \sigma)$   & Succ.$^\dag$
              & Cost $(\mu \pm \sigma)$   & Succ.$^\ddag$ \\\midrule
    MPPI$^\S$ & ---              & ---    & $30.8 \pm 12.6$ & 100\% \\
    \disco    & $250.8 \pm 29.9$ & 20\%   & $61.3 \pm 40.0$ &  70\% \\
    \sv       & $191.7 \pm 56.5$ & 25\%   & $44.5 \pm 17.9$ &  70\% \\
    \dust     & $\mathbf{118.3 \pm 07.9}$ & 100\% 
              & $\mathbf{36.8 \pm 14.0}$  &  80\% \\\bottomrule
    \end{tabular}
    \caption{\textbf{Simulation results}.
    Summary of results for simulation experiments.
    The mean episode cost is given by the sum of the instant costs over the
    episode length.
    Values shown do not include the crash penalty for a more
    comparable baseline.
    $^\S$Not used in the navigation task; has perfect knowledge
    in the pendulum task.
    $^\dag$Successes are episodes with no crashes.
    $^\ddag$Successes are episodes whose last five steps have a instant cost 
    below 4 ($\approx$\ang{10} from the upright position).}
    \label{tab:sim_results}
\end{table}

\subsection{Point-mass navigation on an obstacle grid} \label{sec:exp_part2d}
%%%%%%%%%%%%%%%%%%%%%%%%%%%%%%%%%%%%%%%%%%%%%%%%%%%%%%%%%%%%%%%%%%%%%%%%%%%%%%%%
Here, we reproduce and extend the planar navigation task presented in
\cite{lambert_stein_2020}.
We construct a scenario in which an holonomic point-mass robot must reach a
target location while avoiding obstacles.
As in \cite{lambert_stein_2020}, colliding with obstacles not only incurs a 
high cost penalty to the controller, but prevents all future movement, 
simulating a crash.
The non-differentiable cost function makes this a challenging problem,
well-suited for sampling-based approaches.
Obstacles lie in an equally spaced 4-by-4 grid, yielding several multi-modal
solutions.
This is depicted in \cref{fig:part2d_results}.
Additionally, we include barriers at the boundaries of the simulated space to
prevent the robot from easily circumventing obstacles.

The system dynamics is represented as a double integrator model with 
non-unitary mass $m$, \st the particle acceleration is given by
$\Ddot{\state} = m^{-1} \control$ and the control signal is the force applied
to the point-mass.
In order to demonstrate the inference over dynamics, we forcibly change the mass
of the robot at a fixed step of the experiment, adding extra weight.
This has a direct parallel to several tasks in reality, such as collecting a 
payload or passengers while executing a task.
Assuming the goal position is denoted by $\state_{g}$, the cost function
that defines the task is given by:
\begin{equation*}
\begin{split}
    &\instCost(\state_{\tIdx}, \control_{\tIdx}) =
    0.5 \anyerror_{\tIdx}^\transpose \anyerror_{\tIdx}
    + 0.25 \dot{\state_{\tIdx}}^\transpose \dot{\state_{\tIdx}}
    + 0.2 \control_{\tIdx}^\transpose \control_{\tIdx}
    + p \cdot \Bbbone \{\operatorname{col.}\} \\
    &\termCost(\state_{\tIdx}, \control_{\tIdx}) =
    1000 \anyerror_{\tIdx}^\transpose \anyerror_{\tIdx}
    + 0.1 \dot{\state_{\tIdx}}^\transpose \dot{\state_{\tIdx}} \,,
\end{split}
\end{equation*}
where $\anyerror_{\tIdx} = \state_{\tIdx} - \state_{g}$ is the instantaneous
position error and $p = 10^6$ is the penalty when a collision happens.
A detailed account of the hyper-parameters used in the experiment is presented
in \cref{app:experiments}.

As a baseline, we once more compare against \disco and \sv.
In \cref{fig:part2d_results} we present an overlay of the trajectories for \sv 
and \dust over 20 independent episodes and we choose to omit trajectories of 
\disco for conciseness.
Collisions to obstacles are denoted by a \textbf{x} marker.
Note that in a third of the episodes \sv is unable to avoid obstacles due
to the high model mismatch while \dust is able to avoid collisions by quickly
adjusting to the new model configuration online.
A typical sequential plot of the posterior distribution induced by fitting a GMM
as in \cref{eq:dust_dyn_posterior} is shown on \cref{fig:part2d_ridge_plot}
for one episode.
There is little variation between episodes and the distribution remains stable
in the intermediate steps not depicted. 

\subsection{Trajectory tracking with autonomous ground vehicle}
%%%%%%%%%%%%%%%%%%%%%%%%%%%%%%%%%%%%%%%%%%%%%%%%%%%%%%%%%%%%%%%%%%%%%%%%%%%%%%%%
We now present experimental results with a physical autonomous ground robot
equipped with a skid-steering drive mechanism.
The kinematics of the robot are based on a modified unicycle model, which
accounts for skidding via an additional parameter
\cite{kozlowski_modeling_2004}.
The parameters of interest in this model are the robot's wheel radius
$\wradius$, axial distance $\axdist$, i.e. the distance between the wheels,
and the displacement of the robot's ICR from the robot's centre $\xicr$.
A non-zero value on the latter affects turning by sliding the robot sideways.
The robot is velocity controlled and, although it possess four-wheel drive, 
the controls is restricted to two-degrees of freedom, left and right wheel speed.
Individual wheel speeds are regulated by a low-level proportional-integral
controller.

The robot is equipped with a 2D Hokuyo LIDAR and operates in an indoor environment in our experiments. Prior to the tests, the area is pre-mapped using the gmapping package~\cite{grisetti2007improved} and the robot is localised against this pre-built map.
Similar to the experiment in \cref{sec:exp_part2d}, we simulate
a change in the environment that could be captured by our parametric model 
 of the robot to explain the real trajectories. 
However, we are only applying a relatively simple kinematic model in which
the effects of the dynamics and ground-wheel interactions are not accounted for.
Therefore, friction and mass are not feasible inference choices.
Hence, out of the available parameters, we opted for inferring $\xicr$,
the robot's centre of rotation. 
Since measuring $\xicr$ involves a laborious process, requiring
different weight measurements or many trajectories from the physical hardware
\cite{yi_kinematic_2009}, this also makes the experiment more realistic.
To circumvent the difficulties of ascertaining $\xicr$, we use the posterior
distribution estimated in \cite{barcelos_disco_2020}, and bootstrap our
experiment with $\xicr \sim \normal(0.5, 0.2^2)$.

To reduce the influence of external effects, such as localisation, we defined
a simple control task of following a circular path at a constant
tangential speed.
Costs were set to make the robot follow a circle of \SI{1}{\meter} radius with
$\instCost(\state_\tIdx) = \sqrt{d_{\tIdx}^2+10(s_{\tIdx}-s_0)^2}$, where
$d_\tIdx$ represents the robot's distance to the edge of the circle and
$s_0 = \SI{0.2}{\meter\per\second}$ is a reference linear speed.

The initial particles needed by \dust in \cref{eq:dust_dyn_grad} for the
estimation of $\xicr$ are sampled from the bootstrapping distribution, whereas
for \sv we set $\xicr = %\SI{0.5}{\kilogram\meter\squared}
\SI{0.5}{\meter}
$, 
the distribution
mean.
Again, we want to capture whether our method is capable of adjusting to
environmental changes.
To this end, approximately halfway through the experiment, we add an extra load of approximately \SI{5.3}{\kilogram} at the rear of the robot in order to alter its centre of mass.
These moments are indicated on the trajectories shown in 
\cref{fig:wombot_trajectories}.
In \cref{fig:wombot_costs} we plot the instant costs for a fixed number of steps
before and after the change of mass.
For the complete experiment parameters refer to the \cref{app:experiments}. 

We observe that considering the uncertainty over $\xicr$ and, crucially,
refining our knowledge over it (see \cref{fig:wombot_ridge} in 
\cref{app:experiments}), allows \dust to significantly outperform the \sv
baseline.
Focusing on the trajectories from \sv we note that our estimation of $\xicr$ is 
probably not accurate.
As the cost function emphasises the tangential speed over the cross-track
error, this results in circles correctly centred, but of smaller radius.
Crucially though, the algorithm cannot overcome this poor initial estimation.
\dust initially appears to find the same solution, but quickly adapts,
overshooting the target trajectory and eventually converging to a better result.
This behaviour can be observed both prior to and after the change in the robot's
mass.
Conversely, with the addition of mass, the trajectory of \sv diverged and 
eventually led the robot to a halt.
\section{Conclusions} \label{sec:conclusion}
%%%%%%%%%%%%%%%%%%%%%%%%%%%%%%%%%%%%%%%%%%%%%%%%%%%%%%%%%%%%%%%%%%%%%%%%%%%%%%%%
We present a method capable of simultaneously estimating model parameters
and controls.
The method expands previous results in control as implicit variational inference 
and provides the theoretical framework to formally incorporate uncertainty over
simulator parameters.
By encapsulating the uncertainty on dynamic systems as distributions over a
parametric model, we are able to incorporate domain knowledge and physics
principles while still allowing for a highly representative model. 
Crucially, we perform an \textit{online} refinement step where the agent
leverages system feedback in a sequential way to efficiently update
its beliefs regardless of the size of observation dataset.
% This  simplifies  the  inference  problem  and  significantly reduces the
% number of observations required to characterise the model.

Simulated experiments are presented for a randomised inverted pendulum
environment and obstacle grid with step changes in the dynamical parameters. 
Additionally, a trajectory tracking experiment utilising a custom built AGV
demonstrates the feasibility of the method for online control.
The results illustrate how the simplifications on dynamic inference
are effective and allow for a quick adjustment of the posterior belief
resulting in a \textit{de facto} adaptive controller.
Consider, for instance, the inverted pendulum task.
In such periodic and non-linear system, untangling mass and length can pose
a difficult challenge as there are many plausible
solutions~\cite{ramos_bayessim:_2019}.
Nonetheless, the results obtained are quite encouraging, even with very few
mapping steps per control loop.
% It is important to note that the algorithm is agnostic to the size of the
% observation dataset as we perform sequential online updates to the distribution
% over the parameters of the model. 
% This prevents the need of heuristics to cull to the acquired data over time.

Finally, we demonstrated how, by incorporating uncertainty over the model
parameters, \dust produces more robust policies to dynamically changing
environments, even when the posterior estimation of the model parameters is
rather uncertain.
This can be seen, for instance, in the adjustments made to trajectories in the
obstacle grid.
\newpage
\balance
\printbibliography
\newpage
\newpage \onecolumn
\appendices
\crefalias{section}{appendix}
%%%%%%%%%%%%%%%%%%%%%%%%%%%%%%%%%%%%%%%%%%%%%%%%%%%%%%%%%%%%%%%%%%%%%%%%%%%%%%%%
\section{Algorithm} \label{app:algo}
\begin{algorithm}
\caption{Sim-to-Real in the loop SVMPC}
\label{alg:dust}
\SetAlgoLined
\DontPrintSemicolon

% \KwIn{
% $\pDensity(\simParams)$,
% $\simFunction$,
% $\qDensity(\polParams_{\tIdx_0})$,
% $\controlHorizon$,
% $\samplesSize$\;

% \textbf{Control Hyperparameters}:
% $\temperature$,
% $\controlAuth$,
% $\cost(\cdot)$
% }

Sample $\left\{\polParams_{\tIdx_0}^{\polIdx}\right\}_{\polIdx=1}^{\polSize}
\sim \qDensity(\polParams_{\tIdx_0})$

\lForEach{policy $\polIdx \in \polSize$}{$\policy_{\polParams^\polIdx} \gets
\normal(\polParams_{\tIdx_0}^{\polIdx}, \polPriorCov)$}

\While{task not complete}{
    $\state_{\tIdx}^{r} \gets$ GetStateEstimate()
    
    $\dataset \gets \dataset[\tIdx - 1] \bigcup
    \{\state_{\tIdx}^{r}, \state_{\tIdx-1}^{r}, \control_{\tIdx-1}^{r}$\}
    
    \For(\tcp*[f]{Dynamics inference loop}){$l \gets 1$ \KwTo $\mpfSteps$}{
    \For{$\simIdx \gets 1$ \KwTo $\simSize$}{
        $\likelihood(\simParams \given \dataset) \gets
        \normal(\simFunction(\state^{r}_{\tIdx - 1}, \control_{\tIdx - 1}^{r})
        ; \state^{r}_{\tIdx}, \mpfLikCov)$
        \tcp*{Condition likelihood, \cref{eq:dust_dyn_lik}}
        
        $\nabla_{\simParams}
        \log\pDensity(\simParams^{\simIdx} \given \dataset)
        \approx \nabla_{\simParams}
        \log\likelihood(\simParams^{\simIdx} \given \dataset)
        + \nabla_{\simParams}
        \log\qDensity(\simParams^{\simIdx} \given \dataset[\tIdx-1])$
        \tcp*{\cref{eq:dust_dyn_grad}}
        
        $\scoreFunc(\simParams^{\simIdx}) \leftarrow
        \frac{1}{\simSize} \sum_{\secIdx=1}^{\simSize}
        k(\simParams^{\secIdx}, \simParams^{\simIdx})
        \nabla_{\simParams}
        \log\pDensity_\tIdx(\simParams^{\simIdx} \given \dataset)
        + \nabla_{\simParams}
        k(\simParams^{\secIdx}, \simParams^{\simIdx})$
        \tcp*{Stein gradient, \cref{eq:stein_approx_score_func}}
        
        $\simParams^{\simIdx} \gets \simParams^{\simIdx}
        + \stepSize \scoreFunc(\simParams^{\simIdx})$
        
        $\qDensity(\simParams \given \dataset) = \frac{1}{\simSize}
        \sum_{\simIdx=1}^\simSize \normal(\simParams^\simIdx, \mpfPriorCov)$
        \tcp*{Update posterior, \cref{eq:dust_dyn_posterior}}
    }
    }
    Sample $\left\{\simParams^{\simIdx}\right\}_{\simIdx=1}^{\simSize} 
    \sim \qDensity(\simParams \given \dataset)$
    
    \For(\tcp*[f]{Policies inference loop}){$\polIdx \gets 1$ \KwTo $\polSize$}{
        Sample
        $\left\{\controlSeq_\trajectoryIdx^{\polIdx}\right\}
        _{\trajectoryIdx=1}^{\polSamples} \sim \policy_{\polParams^{\polIdx}}$
        
        \lForEach(\tcp*[f]{\cref{eq:rollout_operator,eq:cost_function}})
        {$\trajectoryIdx \in \polSamples$ and $\simIdx \in \simSamples$} 
        {$\cost_{\trajectoryIdx, \simIdx}^{\polIdx} \gets$
        GetRolloutCosts$(\controlSeq_{\trajectoryIdx}^{\polIdx},
        \simParams^{\simIdx}, \state_{\tIdx}^{r})$}

        $\nabla_{\polParams}
        \log\pDensity(\polParams_{\tIdx}^{\polIdx} \given \optimality, \simParams) 
        \approx \nabla_{\polParams} \log\qDensity(\polParams_{\tIdx-1}^{\polIdx})
        + \nabla_{\polParams} \log\left(\frac{1}{\polSamples\simSamples} 
        \sum_{\trajectoryIdx=1}^{\polSamples}
        \sum_{\simIdx=1}^{\simSamples}
        \exp(-\alpha \cost_{\trajectoryIdx, \simIdx}^{\polIdx})\right)$
        \tcp*{\cref{eq:dust_pol_grad}}
        
        $\scoreFunc(\polParams_{\tIdx}^{\polIdx}) \leftarrow
        \frac{1}{\polSize} \sum_{\secIdx=1}^{\polSize}
        k(\polParams_{\tIdx-1}^{\secIdx}, \polParams_{\tIdx}^{\polIdx})
        \nabla_{\polParams}
        \log\pDensity(\polParams_{\tIdx}^{\secIdx} 
        \given \optimality, \simParams)
        + \nabla_{\polParams}
        k(\polParams_{\tIdx}^{\secIdx}, \polParams_{\tIdx}^{\polIdx})$
        \tcp*{Stein gradient}
        
        $\polParams_{\tIdx}^{\polIdx} \leftarrow \polParams_{\tIdx}^{\polIdx}
        + \stepSize \scoreFunc(\polParams_{\tIdx}^{\polIdx})$
        
        % Re-sample $\left\{\controlSeq_\trajectoryIdx^{\polIdx}\right\}
        % _{\trajectoryIdx=1}^{\polSamples}
        % \sim \normal(\polParams_{\tIdx}^{\polIdx}, \controlAuth)$
        % \tcp*{Get rollouts with updated $\polParams_{\tIdx}$}
        
        % \lForEach{$\trajectoryIdx \in \polSamples$
        % and $\simIdx \in \simSamples$} 
        % {$\cost_{\trajectoryIdx, \simIdx}^{\polIdx} \gets$
        % GetRolloutCosts$(\controlSeq_{\trajectoryIdx}^{\polIdx},
        % \simParams^{\simIdx}, \state_{\tIdx}^{r})$}
        
        $\weight_{\tIdx}^{\polIdx} \leftarrow
        \frac{\qDensity(\polParams_{\tIdx-1}^{\polIdx})}{\polSamples\simSamples} 
        \sum_{\trajectoryIdx=1}^{\polSamples}
        \sum_{\simIdx=1}^{\simSamples}
        \exp(-\alpha \cost_{\trajectoryIdx, \simIdx}^{\polIdx})$
        \tcp*{Compute weights, \cref{eq:dust_action_selection}}
    }
    \lForEach{policy $\polIdx \in \polSize$}
    {$\weight_{\tIdx}^{\polIdx} \leftarrow \frac{\weight_{\tIdx}^{\polIdx}}
    {\sum_{\polIdx=1}^{\polSize} \weight_{\tIdx}^{\polIdx}}$}
    
    $\polIdx^* = \argmax_{\polIdx}{\weight_{\tIdx}^{\polIdx}}$
    
    SendToActuators$(\controlSeq_{\tIdx}^{\polIdx^*} = 
    \polParams_{\tIdx}^{\polIdx^*})$
    \tcp*{Applies first action of the control sequence}
    
    $\polParams_{\tIdx} \gets$ RollPolicies$(\polParams_{\tIdx})$ 
    \tcp*{Shift one step ahead, adds noise to last step}
    
    $\qDensity(\polParams_{\tIdx}) = \sum_{\polIdx=1}^{\polSize} 
    \weight_{\tIdx}^{\polIdx}
    \normal(\polParams_{\tIdx}^{\polIdx}, \controlAuth)$
    \tcp*{Update policies prior, \cref{eq:dust_pol_shift}}
    
    \lForEach{policy $\polIdx \in \polSize$}{$\policy_{\polParams^\polIdx} \gets
    \normal(\polParams_{\tIdx}^{\polIdx}, \polPriorCov)$}
    
    $t \gets t+1$
}
\end{algorithm}

\section{Considerations on Action Selection} \label{app:action_selection}
%%%%%%%%%%%%%%%%%%%%%%%%%%%%%%%%%%%%%%%%%%%%%%%%%%%%%%%%%%%%%%%%%%%%%%%%%%%%%%%%
Once policy parameters have been updated according to \cref{eq:dust_pol_step},
we can update the policy for the current step, $\policy_{\polParams_{\tIdx}}$.
However, defining the updated policy is not enough, as we still need to
determine which immediate control action should be sent to the system.
There are many options which could be considered at this stage.
One alternative would be to take the expected value at each time-step,
although that could compromise the multi-modality of the solution found. 
Other options would be to consider the modes $\policy_{\polParams_{\tIdx}}$ of 
at each horizon step or sample the actions directly.
Finally, we adopt the choice of computing the probabilistic weight of each 
particle and choosing the highest weighted particle as the action sequence for
the current step.
More formally:
\begin{equation} \label{eq:dust_action_selection}
    \weight_{\primIdx} = \frac{
    \pDensity(\optimality \given \polParams_{\tIdx}^{\primIdx}, \simParams) 
    \qDensity(\polParams_{\tIdx-1}^{\primIdx})
    }{
    \sum_{j=1}^{m} 
    \pDensity(\optimality \given \polParams_{\tIdx}^{\secIdx}, \simParams) 
    \qDensity(\polParams_{\tIdx-1}^{\secIdx})
    }
    \approx \pDensity(\polParams_{\tIdx}^{\primIdx} \given \optimality,
    \simParams) \,.
\end{equation}

And finally, the action deployed to the system would be given by
$\controlSeq_{\tIdx}^{\polIdx^*} = \polParams_{\tIdx}^{\polIdx^*}$,
where $\polIdx^* = \argmax_{\polIdx}{\weight_{\tIdx}^{\polIdx}}$.

\section{Considerations on Covariate Shift} \label{app:dyn_evolution}
%%%%%%%%%%%%%%%%%%%%%%%%%%%%%%%%%%%%%%%%%%%%%%%%%%%%%%%%%%%%%%%%%%%%%%%%%%%%%%%%
The proposed inference method assumes that the parameters of the dynamical model
are fixed over time.
Note that, even if the distribution over parameters changes over time, this
remains a plausible consideration, given that the control loop will likely
have a relatively high frequency when compared to the environment covariate
shift.
This also ensures that trajectories generated in simulation are consistent and
that the resulting changes are being governed by the variations in the control
actions and not the environment.

However, it is also clear that the method is intrinsically adaptable to changes 
in the environment, as long as there is a minimum probability of the latent
parameter being feasible under the prior distribution $\qDensity(\simParams 
\given \datapoint)$.
Too see this, consider the case where there is an abrupt change in the 
environment (\eg the agent picks-up some load or the type of terrain changes).
In this situation, $\qDensity(\simParams \given \datapoint[\tIdx - 1])$ would
behave as if a poorly specified prior, meaning that as long the probability density around the true distribution 
$\pDensity(\simParams)$ is non-zero, we would still converge to the true 
distribution, albeit requiring further gradient steps.

In practice, the more data we gather to corroborate a given parameter set, the
more concentrated the distribution would be around a given location in the
parameter space and the longer it would take to transport the probability mass
to other regions of the parameter space.
This could be controlled by including heuristic weight terms to the likelihood 
and prior in \cref{eq:dust_dyn_inf_dataset}.
However, we deliberately choose not to include extra hyper-parameters based on
the hypothesis that the control loop is significantly faster than the changes in
the environment, which in general allows the system to gather a few dozens or 
possibly more observations before converging to a good estimate of 
$\simParams$.

\section{Bias on joint inference of policy and dynamics} \label{app:joint_bias}
%%%%%%%%%%%%%%%%%%%%%%%%%%%%%%%%%%%%%%%%%%%%%%%%%%%%%%%%%%%%%%%%%%%%%%%%%%%%%%%%
Note that, if we take the gradient of \cref{eq:dust_problem} \wrt 
$\simParams$, we get:
\begin{equation}
\begin{split}
    \nabla_{\simParams} \pDensity(\polParams_{\tIdx}, \simParams \given \optimality, \dataset)
    &= \nabla_{\simParams} \left[\pDensity(\polParams_{\tIdx} \given \optimality, \simParams) \pDensity(\simParams \given \dataset)\right] \\
    &= \pDensity(\simParams \given \dataset) \nabla_{\simParams} \pDensity(\polParams_{\tIdx} \given \optimality, \simParams)
    + \pDensity(\polParams_{\tIdx} \given \optimality, \simParams) \nabla_{\simParams} \pDensity(\simParams \given \dataset) \,.
\end{split}
\end{equation}
The first term on the RHS of the equation above indicates that the optimality
gradient of \textit{simulated} trajectories would contribute to the update of
$\simParams$.
Although this is true for the policy updates, we don't want the inference of the
\textit{physical} parameters to be biased by our optimality measure.
In other words, the distribution over $\simParams$ shouldn't conform to whatever
would benefit the optimality policy, but the other way around.

\section{Further details on performed experiments} \label{app:experiments}
%%%%%%%%%%%%%%%%%%%%%%%%%%%%%%%%%%%%%%%%%%%%%%%%%%%%%%%%%%%%%%%%%%%%%%%%%%%%%%%%
\begin{table}[!ht]
    \centering
    \begin{tabular}{lccc} \toprule
    Parameter & Inverted Pendulum & Point-mass Navigation & AGV Traj. Tracking  
    \\\midrule
    Initial state, $\state_0$ &
    [\SI{3}{\radian}, \SI{0}{\meter\per\sec}] & --- & --- \\
    Environment maximum speed &
    \SI{5}{\meter\per\sec} & --- & --- \\
    Environment maximum acceleration &
    \SI{10}{\meter\per\sec\squared} & --- & --- \\
    Policy samples, $\polSamples$ & \num{32} & \num{64} & \num{50} \\
    Dynamics samples, $\simSamples$ & \num{8} & \num{4} & \num{4} \\
    Cost Likelihood inverse temperature, $\alpha$ &
    \num{1.0} & \num{1.0} & \num{1.0} \\
    Control authority, $\controlAuth$ & $2.0^2$ & $5.0^2$ & $0.1^2$ \\
    Control horizon, $\controlHorizon$ & \num{20} & \num{40} & \num{20} \\
    Number of policies, $\polSize$ & \num{3} & \num{6} & \num{2} \\
    Policy Kernel, $k_{\policy}(\cdot, \cdot)$ &
    \multicolumn{3}{c}{Radial Basis Function} \\
    Policy Kernel bandwidth selection & \multicolumn{3}{c}{Silverman's rule}\\
    Policy prior covariance, $\polPriorCov$ & $2.0^2$ & $5.0^2$ & $1.0^2$ \\
    Policy step size, $\stepSize$ & \num{2.0} & \num{100.0} & \num{0.02} \\
    Dynamics prior distribution & \\
    Dynamics number of particles, $\simSize$ & \num{50} & \num{50} & \num{50} \\
    Dynamics Kernel, $k_{\simParams}(\cdot, \cdot)$ &
    \multicolumn{3}{c}{Radial Basis Function} \\
    Dynamics GMM covariance, $\mpfPriorCov$ &
    Improved Sheather Jones & $0.25^2$ & $0.0625^2$ \\
    Dynamics likelihood covariance, $\mpfLikCov$ &
    $0.1^2$ & $0.1^2$ & $0.1^2$ \\
    Dynamics update steps, $\mpfSteps$ & \num{20} & \num{20} & \num{5} \\
    Dynamics step size, $\stepSize$ & \num{0.001} & \num{0.01} & \num{0.05} \\
    Dynamics in log space & No & Yes & No \\
    Unscented Transform spread \cite{barcelos_disco_2020}, $\spread$ &
    \num{0.5} & --- & --- \\
    \bottomrule
    \end{tabular}
    \caption{Hyperparameters used in the experiments.}
    \label{tab:exp_hyperparams}
\end{table}

\end{document}